%% file: main.tex
\documentclass{article}

\PassOptionsToPackage{sort,numbers}{natbib}

\usepackage[preprint]{neurips_2021}

\usepackage[utf8]{inputenc} %
\usepackage[T1]{fontenc}    %
\usepackage{hyperref}       %
\usepackage[svgnames]{xcolor}         %
\hypersetup{
     colorlinks=true,
     linkcolor=NavyBlue,
     filecolor=NavyBlue,
     citecolor=NavyBlue,      
     urlcolor=Magenta,
     }
\usepackage{url}            %
\usepackage{booktabs}       %
\usepackage{amsfonts}       %
\usepackage{nicefrac}       %
\usepackage{microtype}      %
\usepackage{multirow}
\usepackage{xspace}
\usepackage{graphicx}
\usepackage{subcaption}
\usepackage{caption}
\usepackage{amsmath}
\usepackage{wrapfig}
\usepackage{adjustbox}

\usepackage{colortbl} %
\usepackage{soul} %

\usepackage{misc_symbols}

\newcommand{\bert}[0]{{\tt BERT}\xspace}
\newcommand{\distilbert}[0]{{\tt d}\bert}
\newcommand{\roberta}[0]{{\tt RoB}\xspace}
\newcommand{\distilroberta}[0]{{\tt d}\roberta}

\newcommand{\firstinit}[0]{{\tt Init\#1}\xspace}
\newcommand{\secondinit}[0]{{\tt Init\#2}\xspace}
\newcommand{\randinit}[0]{{\tt Untrained}\xspace}

\newcommand{\Testuntrained}[0]{Untrained Model Test\xspace}
\newcommand{\testdiffinit}[0]{different initializations test\xspace}
\newcommand{\Testdiffinit}[0]{Different Initializations Test\xspace}

\newcommand{\AbbrvTestuntrained}[0]{{\tt UMT}\xspace}
\newcommand{\AbbrvTestdiffinit}[0]{{\tt DIT}\xspace}

\DeclareMathOperator*{\argmin}{arg\,min}

\title{More Than Words: Towards Better Quality Interpretations of Text Classifiers}

\author{%
  Muhammad Bilal Zafar,$^{1*}$
  Philipp Schmidt,$^{2*}$
  Michele Donini,$^1$
  \\ 
  \textbf{
  C\'edric Archambeau,$^1$
  Felix Biessmann,$^2$ 
  Sanjiv Ranjan Das,$^{1,3}$ 
  Krishnaram Kenthapadi$^1$
  } \\
  $^1$Amazon Web Services,
  $^2$Amazon Search,
  $^3$Santa Clara University \\
  \texttt{[zafamuh,phschmid,donini,cedrica,biessman,sanjivda,kenthk]@amazon.com} \\
}

\begin{document}

\maketitle

\def\thefootnote{*}\footnotetext{Equal Contribution.}
\def\thefootnote{\arabic{footnote}}

\input{text/abstract}
\input{text/intro}

\input{text/related}

\input{text/setup}

\input{text/sentence_interp}

\input{text/experiments_computational}

\input{text/results_surveys}

\input{text/conclusion}

\bibliographystyle{plain}
\bibliography{NLP_XAI}

\newpage
\input{text/appendix}

\end{document}

%% file: text/abstract.tex
\begin{abstract}
The large size and complex decision mechanisms of state-of-the-art text classifiers make it difficult for humans to understand their predictions, leading to a potential lack of trust by the users. These issues have led to the adoption of methods like SHAP and Integrated Gradients 
to explain classification decisions 
by assigning importance scores to input tokens. However, prior work, using different randomization tests, has shown that interpretations generated by these methods may not be robust. For instance, models making the same predictions on the test set may still lead to different feature importance rankings. In order to 
address the lack of robustness of token-based interpretability, we explore explanations at higher semantic levels like sentences. We use %
computational metrics 
and human subject studies to compare the quality of sentence-based interpretations against token-based ones. Our experiments show that %
higher-level feature attributions offer several advantages: 1) they are more robust as measured by the randomization tests, 2) they lead to lower variability when using approximation-based methods like SHAP, and 3) they are more intelligible to humans in situations where the linguistic coherence resides at a higher granularity level. Based on these findings, we 
show that  token-based interpretability, while being a convenient first choice given the input interfaces of the ML models, is not the most effective one in all situations.
\end{abstract}

%% file: text/intro.tex
\section{Introduction} \label{sec:intro}

Recent advances in natural language processing, especially those aided by large language models like BERT~\citep{devlin_bert_2019}, RoBERTa~\citep{liu_roberta_2019}, GPT-3~\citep{brown_language_2020}, and Switch Transformer~\cite{fedus_switch_2021}, have helped set up new benchmarks for text classification.
While these performance gains have been attributed to the vast amounts of training data and the large number of parameters (i.e., model complexity), it resulted in model predictions being more difficult to interpret.
For text classification tasks, approaches to interpret model predictions are often borrowed from classifiers applied to tabular and image data~\cite{wallace_allennlp_2019}.

As a whole, interpretability for ML models over text data (\eg, Transformers, LSTMs) is an underexplored domain and poses a unique set of challenges. Well-adopted interpretability techniques in the tabular and image domain (\eg, LIME \citep{ribeiro_why_2016}, SHAP \citep{lundberg_unified_2017}, Integrated Gradients \citep{sundararajan_axiomatic_2017}, and feature permutation \citep{breiman_random_2001}) do not provide satisfactory performance on text data as they are often designed to provide interpretability based on individual tokens. 
For instance, recent work has  pointed out that interpretations provided by these methods may suffer from a lack of robustness~\cite{zafar_lack_2021}, where two models with identical architectures and predictions can lead to large differences in interpretations. 
To overcome these issues, we investigate moving away from token-based interpretation and explore interpretations at higher levels of granularity like sentences. 

There are also additional factors that motivate the need to explore higher levels of interpretation granularity.
First, feature importance at the level of tokens is hard to interpret because it removes context surrounding the tokens, leading to a sparse `bag of words' interpretation when the model (\eg, BERT \cite{devlin_bert_2019}) itself is contextually rich. 
This begs the question as to whether coarser than token-level interpretations better accommodate the context. 
Second, good textual interpretations should be pithy and perspicuous~\cite{miller_explanation_2019}. %
Token-based interpretations highlight (too) many tokens, many of which may be duplicates and/or synonymous with each other, %
rendering the semantic information being represented %
possibly confusing. 
Third, token-based interpretations are often not contiguous, resulting in a higher cognitive load for humans. Such increased complexity has been demonstrated to negatively impact the usefulness of interpretations \citep{lage_evaluation_2018} %
as the lack of sequencing may %
prevent us from providing meaningful interpretations. %
Fourth, using sentences instead of tokens as features reduces the size of the feature set, thereby decreasing the computational cost %
of stochastic explainers such as SHAP \citep{lundberg_unified_2017}, and can also help reduce 
the variability in the explainer output. %

\textbf{Contributions.}
We take a first step towards investigating the differences in the quality of token vs. sentence-based interpretations using metrics accounting for the statistical robustness, the computational cost, and the cognitive load for human subjects.
Specifically, (1) we adopt the parameter randomization tests of~\citep{zafar_lack_2021} to assess the robustness of sentence-based interpretations (\S\ref{sec:stability_metric}); (2) we propose metrics to measure the variability in output of stochastic explanation methods like SHAP for text classification, and compare the outcomes for token and sentence-based interpretations (\S\ref{sec:stability_metric}); (3) we design a formal experimental setting to determine which interpretations are better, \ie, to examine whether removal of context matters and which level of granularity lets the humans be most effective (in terms of accuracy and response time) in carrying out an annotation task (\S\ref{sec:human_intel_metrics}).

Our main findings for texts consisting of several sentences suggest the following. 
(1) Sentence-based interpretations exhibit greater robustness as determined by parameter randomization tests~\citep{zafar_lack_2021} and hence are likely to be more trustworthy to users than token-based ones (\S\ref{sec:robustness_tests}).
(2) An analysis of SHAP shows that a coarser level of granularity (sentences) leads to lower variability across multiple runs of the interpretability method on the same input. This low variability can help  mitigate the potential erosion of user trust that can stem from observing different interpretations on the same input (\S\ref{sec:explanation_variance}). 
(3) %
Humans achieve a higher annotation accuracy and a similar or lower response time (indicating lower cognitive load) in predicting the ground truth labels of the text for sentence-based interpretations than for token-based interpretations. 
This improvement in annotation performance suggests that
sentence-based interpretations are significantly more effective than their token-based counterparts (\S\ref{sec:human_intelligibility_results}).

%% file: text/related.tex
\section{Related work}

As mentioned in \S\ref{sec:intro}, most methods for NLP interpretability are borrowed from vision and tabular data domains, see for instance usage of Layerwise Relevance Propagation~\citep{samek_evaluating_2017} in NLP~\cite{poerner_evaluating_2018}.
Also borrowed from vision/tabular are methods like, DeepLIFT \citep{shrikumar_learning_2017}, and LIMSSE (for substring explanations) \citep{ribeiro_why_2016}.

Our work extends the literature on assessment of NLP interpretations \cite{guidotti_survey_2018}.
For instance, \cite{atanasova_diagnostic_2020} finds that gradient-based methods perform the best. 
\cite{lakkaraju_faithful_2019} suggests model agnostic interpretability with accuracy tradeoffs.
Domain specific interpretability has also been assessed for medicine \citep{tjoa_survey_2020} and finance \citep{yang_finbert_2020}.

Current commercial implementations are largely token-based, though they offer excellent visualizations \cite{tenney_language_2020,wallace_allennlp_2019}. Recent  %
concurrent work by \cite{rychener_sentence-based_2020} also advocates for sentence-based NLP interpretability. This analysis is, however, based on a custom-built dataset and does not include human surveys and robustness analysis, though an informal assessment of the quality of interpretations is undertaken by the authors. \cite{gilpin_explaining_2018} draws a distinction between interpretability (feature importance based on system internals) versus explanations (human ingestible answer to a question) and suggests that sentence-based explanations may offer more than mere interpretability. This is related to arguments in \cite{bhatt_explainable_2020}, suggesting that current interpretability approaches serve internal users more than external ones, as they lack transparency, an aspect that  sentence-based explanations may improve.  \cite{ribeiro_beyond_2020} is a recent attempt at human-in-the-loop (HITL) development of NLP models; they introduce a task-agnostic checklist approach which also generates diverse test cases to detect potential bugs. Here the human involvement is in generating test cases, not in evaluating the quality of interpretations.
Some of our questions are similar to those assessed by \cite{lundberg_consistent_2019} for tree models, and they demonstrate  improvements in run time, clustering performance, identification of important features, assessed via a user study. 
The recently proposed XRAI approach~\cite{kapishnikov_xrai_2019} also advocates presenting interpretations by grouping multiple features, though unlike us, they focus on the image domain.

How the interpretation is presented to the human subject matters. \cite{lage_evaluation_2018} demonstrates that the length and complexity of interpretations have significant impact on the effectiveness as measured in HITL metrics. 
Highlighting important tokens is a popular way of presenting explanations. An alternative is the removal of non-important tokens.
However, comprehensibility of the interpretation drops considerably with this approach  \citep{feng_pathologies_2018}. Borrowing visualization methods from computer vision (\eg, \cite{krizhevsky_imagenet_2012,adebayo_sanity_2018}),  \cite{li_visualizing_2016} explores how negation, intensification \etc{} may be built up into sentence-based salience. 

Text interpretations may be improved using anchors \citep{ribeiro_anchors_2018}, where rules are encoded as super features that override other features, for example negations (\eg, good vs \emph{not} good). Anchors may be a way to get highly parsimonious interpretations, but may not be sufficiently explanatory for long texts. Phrase-based interpretations such as topical n-grams may be more fruitful than using tokens for unsupervised tasks~\cite{wang_topical_2007}. 
\cite{arras_evaluating_2019} explores the relevance of token-based feature importance to build up sentence-based interpretations. 
There are related questions as to whether pre-trained language models (\eg, \cite{beltagy_scibert_2019}) that are domain specific lead to better interpretations than generic ones, such as BERT \citep{devlin_bert_2019} or RoBERTa \citep{liu_roberta_2019}. Using sentence-based embeddings from BERT, such as SBERT \citep{reimers_sentence-bert_2019}, may also generate better interpretations. Our work complements all these varied experiments in an attempt to understand better what form of interpretations would be most \textit{comprehensible to human subjects}, which after all, is the goal of machine learning interpretability.  

Finally, a lack of interpretability is not the only risk associated with complex text models. Recent work has highlighted several other issues like environmental impact and bias~\citep{bender_dangers_2021,strubell_energy_2019,dhamala_bold_2021,abid_persistent_2021}. Reliable interpretability methods can be used to detect bias in model behavior~\citep{lipton_mythos_2018,doshi-velez_towards_2017}.

%% file: text/setup.tex
\section{Background on token-based interpretability}
\label{sec:background}

In this section, we briefly describe the functionality of two popular token-based interpretability methods, SHAP~\cite{lundberg_unified_2017} and Integrated Gradients (IG)~\cite{sundararajan_axiomatic_2017}. 
We picked these methods as 
(i) they are shown to provide superior empirical performance as compared to their counterparts like LIME~\cite{ribeiro_why_2016} (see for instance~\cite{lundberg_unified_2017}),
(ii) provide desirable axiomatic properties~\cite{lundberg_unified_2017,sundararajan_axiomatic_2017,sundararajan_many_2020}, and
(iii) lend themselves readily to interpretability at meta-feature level (\eg, phrases, sentences) as we will discuss later in \S\ref{sec:metafeature-interp}.

We assume a tokenized input text, $\tb = [t_1, \ldots, t_T]$, where $T$ is the number of tokens. %
The task is to classify the input into one of $K$ classes.
In this work, we focus on Transformer-based neural text classifiers and hence assume that the tokenization is done using the tokenizer accompanying the model (\eg, WordPiece for BERT, BPE for RoBERTa/GPT-2).
For neural text models, %
the classifier produces a score for each class, $F(\tb) \in \RR^K$. The input is then assigned the label of the class with the highest score. 
We refer to the score corresponding to this class as $f(\tb) \in \RR$.

Given the model score for the predicted class $f(\tb)$, the task of the token-based interpretability methods is to obtain a vector $\mathbf{\Phi}(\mathbf{t}) = [\phi(t_1), \ldots, \phi(t_T)]$ of token attributions that assign an importance score to each of the tokens. The score $\phi(t_i)$ (for brevity, also denoted as $\phi_i$) indicates the importance of token $t_i$ in predicting the score $f(\tb)$.

SHAP, which is a model-agnostic method, estimates the importance of a token by simulating its absence from different subsets $\cb$ of tokens, called coalitions, with $\cb \subseteq \tb$, and computing the average marginal gain of adding the token in question to these. %
Concretely, the token importance is obtained by solving %
the following optimization problem:
\begin{align} \label{eq:shap_token}
    \Phi = \argmin_{\Phi} \sum_{\cb \subseteq \tb} [f_{\cb}(\cb)  - (\phi_0 + \sum_{w\in \cb}\phi(w))]^2 \times k_{SHAP}(\tb, \cb)
\end{align}
where $\tb$ is the original input, $\cb$ is a sub-part of the input $\tb$ corresponding to the tokens that are not dropped, $w$ are the tokens contained within $\cb$, $k_{SHAP}(\tb, \cb)$ represents the SHAP kernel, $\phi_0$ refers to the model output on an empty input (all tokens dropped), and $f_{\cb}$ refers to the model output on the remaining (non-dropped) tokens. 
We simulate token dropping by replacing the corresponding tokens with the unknown vocabulary token. 

The exact computation requires solving Eq.~\ref{eq:shap_token} over all the $2^{T}$ subsets of tokens.
The exact computation becomes infeasible for even modestly large number of input features (\eg, 30). In practice, a much smaller number of coalitions is used. For instance, the SHAP library suggests a heuristic of using $2^{|\tb|} + 2048$ coalitions~\cite{lundberg_shap_2018}. Some recent studies aim at using the output uncertainty~\cite{slack_reliable_2021} or variance~\cite{covert_improving_2021} to define a tradeoff between accuracy and number of coalitions.

Integrated Gradients (IG), which is a model-specific method (in that it requires access to model gradients), operates by estimating the token attribution as:

\begin{align} \label{eq:ig}
     \Phi = (\xb - \bar{\xb}) \odot \int_0^1  \
		            \frac{\partial f(\bar{\xb} + \alpha(\xb-\bar{\xb}))}{\partial \xb} d\alpha,
\end{align}
where $\xb_i$ represents the embedding of input token $t_i$,
$\odot$ is the Hadamard product,
and $\bar{\xb}_i$ 
is the embedding of the baseline token (\eg, unknown vocabulary token).
In a manner similar to SHAP computation, the integral is empirically estimated using summation over the line connecting $\xb$ and $\bar{\xb}$. In a manner similar to SHAP, a sum over larger number of terms leads to better approximation~\cite{sundararajan_axiomatic_2017}.

By construction, both, SHAP and IG have the desirable property that the sum of token attributions equals the predicted score of the class, that is $\sum_{i=0}^{T} \phi_i = f(\tb)$. 
Here, $\phi_0$ is the model output on the baseline input (\ie, no features present), which corresponds to computing $f_{\emptyset}(\emptyset)$ for SHAP and $f(\bar{\xb})$ for IG.

%% file: text/sentence_interp.tex
\section{Interpretability based on meta-tokens}
\label{sec:metafeature-interp}

We describe how to generate interpretations based on meta-tokens. While meta-tokens  can be constructed at various granularities (\eg, phrases, paragraphs), \textit{in this paper, we limit ourselves to sentences}. We also describe the setup for comparing the quality of token-based and sentence-based interpretations.

\subsection{Generating meta-token interpretations}
\label{sec:generating_metatoken}

We assume that the tokenized text $\tb = [t_1, \ldots, t_T]$ can be partitioned into non-overlapping meta-tokens $\bm{m} = [m_1, \ldots, m_M]$ where each meta-token consists of one or more contiguous tokens. 
When meta-tokens are sentences, the partitioning can be done using off-the-shelf sentencizers like spacy~\cite{honnibal_spacy_2020}.

Given token-based attributions $\mathbf{\Phi}(\mathbf{t}) = [\phi(t_1), \ldots, \phi(t_T)]$, one can form SHAP and IG meta-token  %
attributions by summing the attributions of all the tokens corresponding to a meta-token $\bm{m}$. This follows from the summation property of these methods mentioned in \S\ref{sec:background} and the commonly taken feature-independence~\cite{lundberg_unified_2017,sundararajan_many_2020} / marginal expectation~\cite{janzing_causal_2019} assumption. We refer to this procedure for computing meta-token %
attributions as the \textit{indirect} method. 
However, for SHAP, one can also \textit{directly} compute the feature attributions based on meta-tokens
by directly solving Eq.(\ref{eq:shap_token}) where the feature coalitions are formed by dropping full meta-tokens instead of individual tokens. This variant where feature coalitions are formed based on meta-tokens %
will reduce the number of all possible coalitions as $M \ll T \implies 2^M \ll 2^T$, and as a result, can lead to more accurate estimations for the same computational budget.

\subsection{Computational evaluation metrics}
\label{sec:stability_metric}

To compare token vs. sentence-based attributions, we first extend the randomization tests of \cite{zafar_lack_2021}.
Then, we propose a metric to compare the variability of interpretations for stochastic methods like SHAP.

\xhdr{Metrics based on parameter randomization~\citep{zafar_lack_2021}}
Given the training data and an interpretability method, \cite{zafar_lack_2021} start by training three different models: (i) a fully trained model referred to as \firstinit, (ii) another fully trained model \secondinit that is identical to \firstinit in all aspects (\eg, training data, training procedure like batching, and learning rate) except for the initial randomized parameters, and (iii) an untrained model \randinit which is obtained by taking \firstinit and randomizing the weights of all layers except the encoder ones. Given these models, the following tests are carried out:

\textit{\Testdiffinit (\AbbrvTestdiffinit).} This test measures the overlap in feature attributions of the same input from two functionally equivalent models \firstinit and \secondinit (\ie, models that agree on all of the test inputs~\cite{sundararajan_axiomatic_2017,zafar_lack_2021}). 
The overlap in the attributions is measured using Jaccard@K\% metric~\cite{zafar_lack_2021}  (also called the intersection-over-union metric in~\cite{deyoung_eraser_2020}) between the two ranked attribution lists. Given the two sets of attributions $\Phi_i, \Phi_j$ for the same input, Jaccard@K\% is defined as $J(\Phi_i, \Phi_j) = \frac{|\rb_i \cap \rb_j|}{|\rb_i \cup \rb_j|}$ where $\rb_i$ is the set of top-K\% tokens as specified according to the attribution $\Phi_i$.

\textit{\Testuntrained (\AbbrvTestuntrained).} This test measures the overlap in feature attributions of the same input between the trained model \firstinit and the untrained model \randinit. 

The authors in \cite{zafar_lack_2021} argue that a low
overlap according to \AbbrvTestdiffinit or a high
overlap according to \AbbrvTestuntrained %
suggests 
an interpretability method is not 
robust.

To compare the quality of token- and sentece based interpretations,  we conduct the tests separately for both token- and sentence-based interpretations. For each test, let $J_{t}$ denote Jaccard@K\% under token-based interpretations and $J_{s}$ denote Jaccard@K\% under sentence-based interpretations. We analyze the quantity $J_s - J_t$ to compare the robustness. 
$J_s - J_t > 0$ for \AbbrvTestdiffinit implies greater robustness for sentences. Similarly, $J_s - J_t < 0$ for \AbbrvTestuntrained implies greater robustness for sentences.

\xhdr{Overlap.}
We also propose a second measure to test the variability of the interpretations.
Specifically, inspired by the variance analysis of \cite{covert_improving_2021}, we measure the extent to which the top-ranked interpretations vary across \textit{different runs} of the interpretability method with the \textit{same model on the same input}.
Given a model and an input $\tb$, we compute the feature attributions $L$ times. For $1 \le j \le L$, let $\Phi^{j}$ denote the outcome of $j^{th}$ run.
Then, we compute the median pairwise overlap
(see \S\ref{sec:robustness_tests} for reasoning on the choice of median over mean)
between attributions obtained from different runs of the attribution method as:
\begin{align}
\label{eq:overlap}
    \text{overlap} = \text{median} (\{J(\Phi^i, \Phi^j) | 1 \le i < j \le L\}).
\end{align}
A high overlap means higher commonality between the top-K\% attributions across different runs of the interpretability method.

\subsection{Human intelligibility measures} 
\label{sec:human_intel_metrics}

A crucial goal of explainable AI (XAI) methods is to render ML predictions more intelligible for human observers. The extent to which robustness measures (or more generally XAI quality metrics without humans in the loop) correlate with human perception is the subject of active research \citep{lage_evaluation_2018, poursabzi-sangdeh_manipulating_2021, hase_evaluating_2020}. While studies with human subjects are a prerequisite for effective evaluation of XAI methods, there is no single human-in-the-loop metric that would account for all relevant aspects of XAI quality. Two metrics have become popular in the context of XAI quality assessment: (a) simulatability \citep{lipton_mythos_2018}, that is, using the features identified as important by the XAI method, how well humans can replicate ML predictions, and (b) annotation quality, meaning how accurately human annotators can replicate the ground truth label using the important features. Both metrics have been used to assess the impact of transparency in ML models on human-ML interaction \citep{lage_evaluation_2018, hase_evaluating_2020}.  

For human studies in this paper, we chose the annotation quality metric. This choice is motivated by the following two reasons:
First, %
the models  %
obtain close to perfect test accuracy
on the datasets used in the surveys (Appendix~\ref{app:training_results}), hence simulatability is almost equivalent to  annotating the ground truth. Second, and  more importantly, the impact of transparency on ground truth annotations can be considered as more relevant for real world scenarios: in many XAI applications, ML predictions are used to assist humans in annotating ground truth and explanations are used to validate ML predictions \citep{lage_evaluation_2018}. To measure the usefulness of interpretability in such assistive settings we ask subjects to annotate the ground truth. In contrast to simulatability, this metric allows to assess cases when human annotations, ML predictions, and ground truth are not the same, for instance when explanations bias human annotations to blindly follow wrong predictions or in scenarios where ML models need to be debugged.

In our experiments we randomly assign subjects to one out of three experimental conditions: (1) \textit{control}, where no XAI-based highlights were shown to users, (2) \textit{token}, where individual words were highlighted as part of the task, and (3) \textit{sentence}, where entire sentences were highlighted. In both token and sentence cases, we highlight top-10\% features, that is, highlight tokens / sentences, starting from ones with the highest attribution, until 10\% of the tokens in the text have been highlighted.

As part of our evaluation, we report on (1) human accuracy for ground truth annotation and (2) time taken for annotations.  Following~\cite{schmidt_quantifying_2019}, we combine (1) and (2) using the Information Transfer Rate metric $ITR = \frac{I(y_h, y)}{t}$, where $y$ denotes the true label, $y_h$ is the label annotated by the human, $t$ is the average response time, and $I$ is the mutual information. An interpretability method that is more helpful for humans will have a higher annotation throughput, measured by ITR in bits/s.

%% file: text/experiments_computational.tex
\section{Experiments with computational measures}
\label{sec:exp_computational}

We now describe the experiments comparing token-based and sentence-based interpretability. 

\textbf{Datasets.}
With the goal of covering different application domains and data characteristics, we consider 
the following 
three %
datasets. %
\textit{IMDB}: The movie review data where the task is to assess the sentiment (positive or negative) of a movie review.
\textit{Medical}: Medical text data, where the task is to classify the condition of a patient into one of five classes from a medical abstract.
\textit{Wiki}: The Wikipedia article data where the task is to predict whether an article is written with a promotional or neutral tone. 
Appendix~\ref{app:datasets} provides more details on the datasets, their sources, and licenses.

\textbf{Models, training, and hyperparameters.}
Following  \cite{zafar_lack_2021},
we focus on pretrained Transformers, specifically, BERT (\bert), RoBERTa (\roberta), and their distilled versions namely DistilBERT (\distilbert) and DistilRoBERTa (\distilroberta). All experiments were performed on AWS g4dn.xlarge instances. The architectures, training details, and the information on the software used are included in Appendix~\ref{app:reproducability}.

For both tokens and sentences, following the author implementation~\citep{lundberg_shap_2018}, we set the number of coalitions for the SHAP computation to $2M + 2^{11}$, where $M$ is the number of features (\eg, tokens or sentences). For IG, following the original paper~\cite{sundararajan_axiomatic_2017}, we use  $300$ as the number of iterations. Due to the high computational cost of obtaining SHAP attributions for large models, we limit the analysis of attribution comparison to $1,000$ randomly selected test inputs.
For the computation of Jaccard@K\%, following~\citep{zafar_lack_2021}, we set the value of K to 25\% as \citep{zafar_lack_2021} obtained similar results for other values.
The model accuracy and overlap statistics between different initializations are shown in Appendix~\ref{app:training_results}.

We now describe the results comparing the token vs. sentence-based interpretations using the metrics described in \S\ref{sec:metafeature-interp}.
For expositional simplicity, the results reported in this section are with the SHAP algorithm, and sentence-based attributions are computed using the \textit{direct} method in \S\ref{sec:metafeature-interp}. The results for IG method and SHAP with with indirect computation are briefly summarized at the end of this section, and the detailed results can be found in Appendix~\ref{app:sentence_indirect}.

\begin{figure}
     \centering
     \begin{subfigure}[b]{0.45\textwidth}
         \centering
        \includegraphics[width=\textwidth]{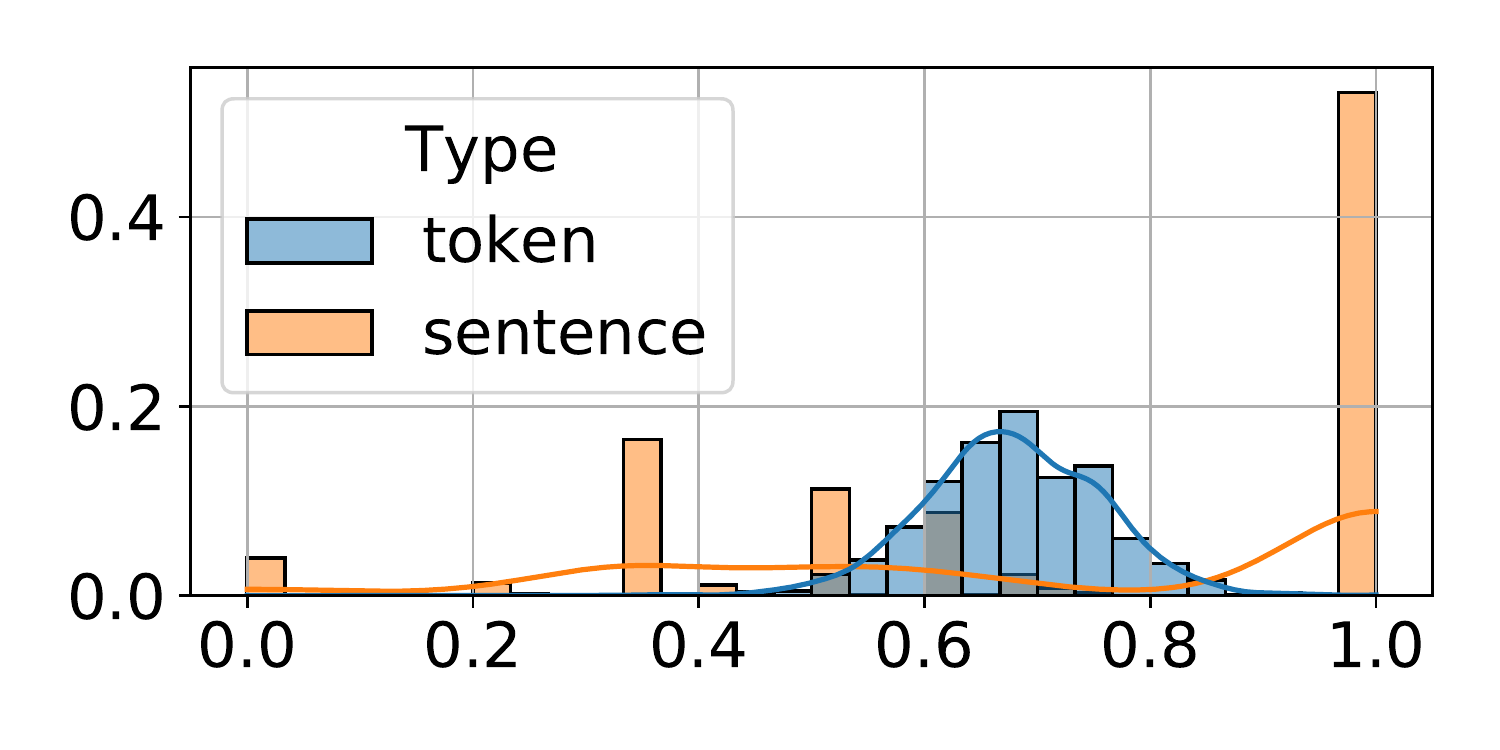}
         \caption{Sentence based interpretations are more robust w.r.t. \AbbrvTestdiffinit. With sentence-based interpretations, over $50\%$ of the cases result in a Jaccard@25\% value of $1$ meaning that the top $25\%$ interpretations for two functionally equivalent models (\firstinit vs. \secondinit) are identical.}
         \label{fig:imdb_jac_first_vs_second}
     \end{subfigure}
     \hfill
     \begin{subfigure}[b]{0.45\textwidth}
         \centering
        \includegraphics[width=\textwidth]{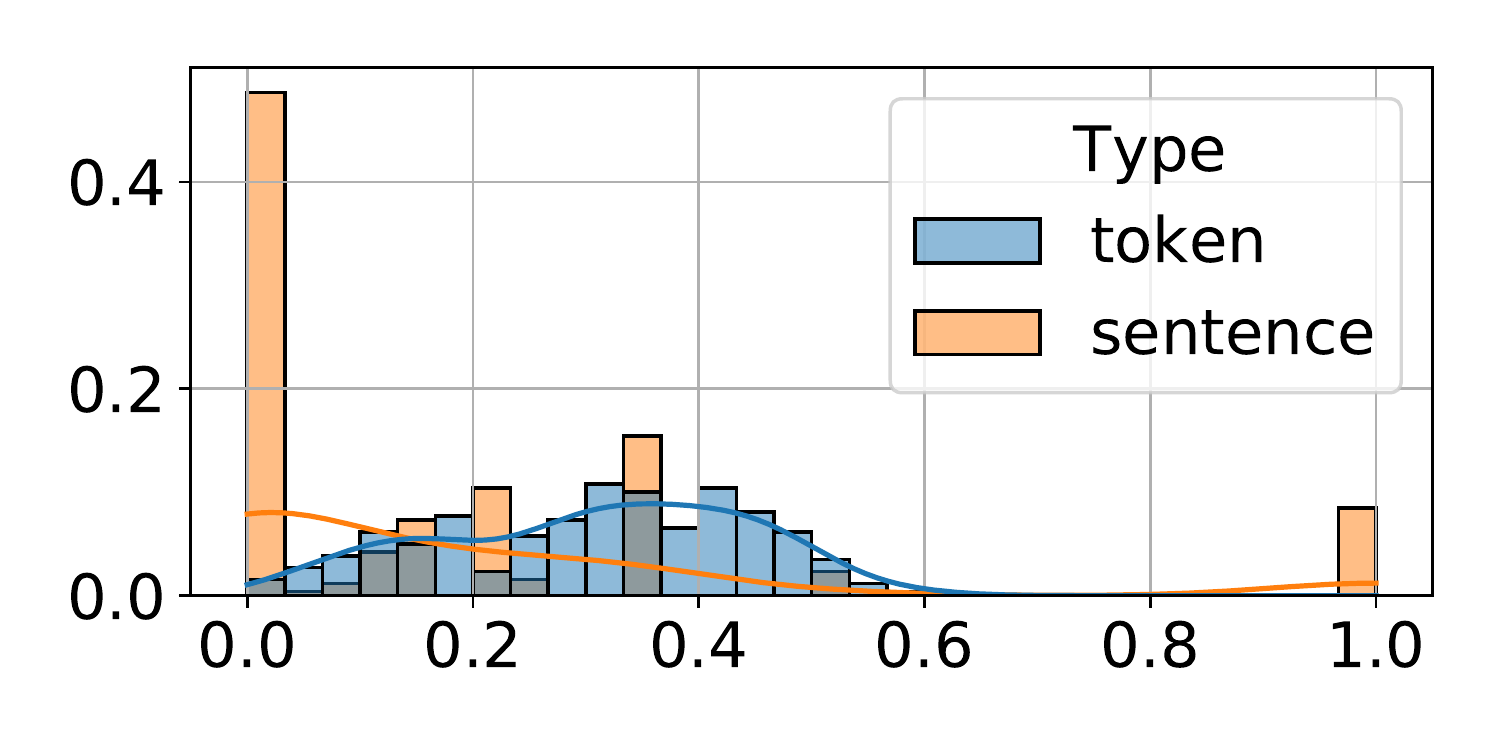}
         \caption{Sentence-based interpretations are more robust w.r.t. \AbbrvTestuntrained. With sentence-based interpretations, almost $50\%$ of the cases result in a Jaccard@25\% value of $0$ meaning that the top $25\%$ interpretations for trained and untrained models (\firstinit vs. \randinit) have no overlap.}
         \label{fig:imdb_jac_first_vs_untrained}
     \end{subfigure}
        \caption{Histogram of Jaccard@25\% with a \bert{} model on IMDB data. Higher values for sentences with \AbbrvTestdiffinit (than tokens) and much lower values with \AbbrvTestuntrained suggest that sentences provide more robust interpretations.}
        \label{fig:imdb_jac}
\end{figure}

\subsection{Robustness under the randomization tests}
\label{sec:robustness_tests}

Figure~\ref{fig:imdb_jac} shows the result of the two randomization tests with SHAP attribution on the IMDB dataset when trained with \bert model. 
Specifically, Figure~\ref{fig:imdb_jac_first_vs_second} shows the histogram of similarity
between interpretations generated using two functionally equivalent models, whereas Figure~\ref{fig:imdb_jac_first_vs_untrained} shows the similarity between trained and untrained models.
In Figure~\ref{fig:imdb_jac_first_vs_second}, we notice that in over half of the cases ($53\%$), Jaccard@25\% is $1$, that is, the set of top-25\% sentences (ranked according to their attributions) are identical between the two functionally equivalent models (\firstinit and \secondinit). On the other hand, Jaccard@25\% is never $1.0$ for tokens.
So as expected, sentences lead to higher robustness w.r.t. the \testdiffinit.
In Figure~\ref{fig:imdb_jac_first_vs_untrained}, we notice that in almost half of the cases ($49\%$), Jaccard@25\% is $0$ when comparing sentence-based interpretations between \firstinit and \randinit. In other words, the sets of top-25\% sentences have no overlap between the trained model (\firstinit) and the untrained model (\randinit). On the other hand, Jaccard@25\% is $0$ only $8\%$ of the times for tokens. Again, sentence-based interpretations are more robust.

\input{tables/shap_direct_jaccard_difference}

We present the results over all the models and datasets in  Table~\ref{table:shap_direct_jac_diff}. The results show the median
difference in Jaccard@25\%  between sentences and tokens for the two tests. 
We compare the medians instead of the means due to the asymmetric and long-tailed nature of the distributions of Jaccard@25\% for sentences in Figure~\ref{fig:imdb_jac}.
The table shows that sentences have a higher median Jaccard@25\% than tokens for IMDB and Medical dataset when comparing \firstinit and \secondinit (Table~\ref{table:shap_direct_jac_diff_init1_init2}). The trend is however reversed for the Wiki dataset where the tokens are more robust. When comparing \firstinit and \randinit, sentences have lower median Jaccard@25\% in 10 out of 12 cases  (Table~\ref{table:shap_direct_jac_diff_init1_untrained}). The only cases where tokens are more robust is the \roberta model on IMDB and Medical datasets.
Appendix~\ref{app:sentence_indirect} shows the results for indirect computation of sentence attribution using SHAP (\S\ref{sec:metafeature-interp}). The results show a very similar trend as in Table~\ref{table:shap_direct_jac_diff}. The corresponding results for IG (Appendix~\ref{app:sentence_indirect}) however show that while sentence-based interpretations are more robust w.r.t. \AbbrvTestdiffinit in most cases, they are less robust w.r.t. \AbbrvTestuntrained.

To summarize, the results show that sentences lead to a higher robustness w.r.t. both randomization tests when using SHAP, whereas for IG, sentences are more robust only w.r.t. \AbbrvTestdiffinit.

\input{tables/shap_direct_multiruns_jaccard}

\subsection{Interpretation variability}
\label{sec:explanation_variance}

For each input, we generate the interpretations 10 times (that is, $L=10$ from \S\ref{sec:stability_metric}). 
Following the description in \S\ref{sec:stability_metric}, we present the median value of the overlap metric for both token and sentence-based interpretations in Table~\ref{table:shap_direct_jac_multiruns}.
Table~\ref{table:shap_direct_jac_multiruns_trained} shows that in  all  configurations, the overlap is %
positive, implying that sentences lead to higher overlap across different runs of the interpretability method. The same holds for the untrained model as well (Table~\ref{table:shap_direct_jac_multiruns_untrained}).
However, we notice that with indirect computations of sentence-based SHAP attributions (Table~\ref{table:shap_indirect_jac_multiruns} in Appendix~\ref{app:sentence_indirect}), sentences often lead to higher variability than tokens.

%% file: tables/shap_direct_jaccard_difference.tex
\begin{table}[t]
    \caption{Positive (median) values of $J_s - J_t$ (\S\ref{sec:stability_metric}) for \AbbrvTestdiffinit in 8/12, and negative values for \AbbrvTestuntrained in 10/12 cases mean that sentence-based interpretations are more robust. }
    \label{table:shap_direct_jac_diff}
    \begin{subtable}{.5\linewidth}
      \caption{\AbbrvTestdiffinit: Median difference in Jaccard@25\% of sentences and tokens when comparing over \firstinit and \secondinit. Positive values mean sentences are more robust.}
      \centering
        \begin{tabular}{lrrrr}
        \toprule
        {} &  \bert &  \roberta &  \distilbert &  \distilroberta \\
        \midrule
        \textbf{IMDB   } &     20 &         5 &           14 &              22 \\
        \textbf{Medical} &     15 &        11 &            9 &              17 \\
        \textbf{Wiki   } &    -25 &       -19 &          -25 &              -4 \\
        \bottomrule
        \end{tabular}
        \label{table:shap_direct_jac_diff_init1_init2}
    \end{subtable}%
    \quad
    \begin{subtable}{.5\linewidth}
      \centering
    \caption{\AbbrvTestuntrained: Median difference in Jaccard@25\% of sentences and tokens when comparing over \firstinit and \randinit. Negative values mean sentences are more robust.}
    \begin{tabular}{lrrrr}
        \toprule
        {} &  \bert &  \roberta &  \distilbert &  \distilroberta \\
        \midrule
        \textbf{IMDB   } &    -17 &        14 &          -16 &             -11 \\
        \textbf{Medical} &     -5 &         2 &          -14 &             -17 \\
        \textbf{Wiki   } &    -23 &       -16 &          -44 &              -6 \\

        \bottomrule
        \end{tabular}
        \label{table:shap_direct_jac_diff_init1_untrained}
    \end{subtable} 
\end{table}

%% file: tables/shap_direct_multiruns_jaccard.tex
\begin{table}[t]
    \caption{Median difference between overlap (\S\ref{sec:stability_metric}) of sentences and tokens when computing the feature attributions repeatedly 10 times. A positive value means that sentences-based attributions are more stable.}
    \label{table:shap_direct_jac_multiruns}
    \begin{subtable}{.5\linewidth}
      \caption{Trained model (\firstinit)}
      \centering
        \begin{tabular}{lrrrr}
        \toprule
        {} &  \bert &  \roberta &  \distilbert &  \distilroberta \\
        \midrule
        \textbf{IMDB   } &                 62 &            67 &                       65 &                  64 \\
        \textbf{Medical} &                 65 &            25 &                       63 &                  63 \\
        \textbf{Wiki   } &                 12 &            35 &                       36 &                  36 \\
        \bottomrule
        \end{tabular}
        \label{table:shap_direct_jac_multiruns_trained}
    \end{subtable}%
    \quad
    \begin{subtable}{.5\linewidth}
      \centering
    \caption{Untrained model (\randinit)}
    \begin{tabular}{lrrrr}
        \toprule
        {} &  \bert &  \roberta &  \distilbert &  \distilroberta \\
        \midrule
        \textbf{IMDB   } &                 19 &            66 &                       29 &                  66 \\
        \textbf{Medical} &                 64 &            64 &                       63 &                  62 \\
        \textbf{Wiki   } &                 36 &            34 &                       19 &                  33 \\
        \bottomrule
        \end{tabular}
        \label{table:shap_direct_jac_multiruns_untrained}
    \end{subtable} 
\end{table}

%% file: text/results_surveys.tex
\section{Impact of interpretation granularity on human annotation performance}
\label{sec:human_intelligibility_results}

We use the \bert model, and  IMDB and Wiki datasets. We omit the Medical dataset as the highly technical medical terminology would require access to human subjects who are experts in the domain.

For each dataset, we randomly assigned human judges to one out of three conditions: 
control, token and sentence (detailed description in \S\ref{sec:human_intel_metrics}).
We chose sample sizes according to a power pre-registration: We determined the minimum sample count for our studies by specifying Type I ($\alpha = 0.05$) and Type II ($\beta = 0.2$) error rates, with estimated effect sizes coming from a pilot study. To execute the user surveys, we carefully compiled annotation instructions that explained the task as well as the payment details to the annotators. We ran all user studies on Toloka, 
a paid site for crowdsourced data labeling.\footnote{\url{toloka.yandex.com}}  
As described in \S\ref{sec:human_intel_metrics}, for both \textit{token} and \textit{sentence} conditions, we highlighted $10\%$ of the text. 
Further details on the human surveys including the data pre-processing, the screenshots of the survey, payment information and quality control can be found in Appendix~\ref{app:surveys}.

\input{tables/survey_gt}

For \textbf{IMDB}, we collected a total of $N = 2150$ samples across the three conditions. 
In Table~\ref{table:imdb_summary}, 
we show human performances in terms of their ability to annotate the ground truth correctly and their task times. 
We note, that both the accuracy and annotation time deteriorate with interpretations (\textit{token} and \textit{sentence}), however, \textit{sentence} still perform better than \textit{token}.
The relation between treatments and human performance was significant, $\chi^2 (2, N = 2150) = 14.15,\,p = 0.001$. A pairwise comparison between \textit{token} and \textit{sentence} condition with Bonferroni-adjusted significance level ($\alpha = 0.017$) 
was above the significance threshold, $\chi^2 (1, N = 1400) = 4.65,\,p = 0.03$.

For the \textbf{Wiki} dataset, we collected a total of $N = 1950$ samples. In Table~\ref{table:wiki_summary}, conditions \textit{control} and \textit{token} have comparable human annotation accuracy, however for \textit{token}, the annotation time increases significantly, indicating increased cognitive load. Sentence-level explanations, in contrast, lead to substantially higher annotation accuracy and less time spent on tasks.
A comparison of ITR between tokens and sentences shows that \textit{sentences increase the annotation throughput (measured in bits/s) by over 150\%}. We conducted a chi-square test of independence to examine the relation between treatments and human performance. The relation between conditions and human performance was not significant, $\chi^2 (2, N = 1950) = 3.37,\,p = 0.19$. A pairwise comparison with Bonferroni-adjusted significance level ($\alpha = 0.017$) between \textit{token} and \textit{sentence} was above the significance threshold, $\chi^2 (1, N = 1350) = 2.68,\,p = 0.10$.

While sentences perform better than tokens in both datasets, the inclusion of sentence highlights deteriorates the annotation performance in the IMDB data when compared to the control condition. We hypothesize that this discrepancy may be due to the difference in the \textit{difficulty level} of the tasks: While sentences do not improve the annotation performance on the easy IMDB data (accuracy of $0.87$ with control), they increased it significantly on a more difficult Wiki data (accuracy of $0.62$ with control). 
To summarize, we notice that \textit{sentence-based interpretations lead to significantly better human intelligibility (measured via response time and accuracy) as compared to token-based ones}.

Finally, we compare the human annotation performance with a closely related computational metric of \textit{infidelity} in Appendix~\ref{app:infidelity}. Our analysis shows a possible misalignment between the two metrics.

%% file: tables/survey_gt.tex
\begin{table}[t]
    \caption{
    Survey results on the IMDB and Wiki datasets, showing the human accuracy in predicting the ground truth under the control (C), token (T), and sentence (S) conditions. Also shown are the aggregated time taken by the participants and the Information Transfer Rate (ITR). In both cases, sentences (S) provide higher ITR than tokens (T). Best condition in \textbf{boldface} per column.
    }
    \label{table:survey}
    \begin{subtable}{.47\linewidth}
        \caption{
        Both token and sentence-based interpretations decrease human annotation accuracy as compared to control with similar cognitive load for annotators, as indicated by the task response times. Combining accuracy and time via ITR shows that both tokens and sentences lead to a lower ITR than control, but \textit{sentences perform better than tokens}. 
        }
        \label{table:imdb_summary}
        \centering
        \begin{adjustbox}{max width=\textwidth}
        \begin{tabular}{llll}
        \toprule
        {} & Accuracy &    Time [s] &    ITR [bits/s]  \\
        \midrule
        \textit{C}   &
        \textbf{0.87 ± 0.33} &   2111.11 ± 527.99 &  \textbf{0.009 ± 0.003}  \\
        \textit{T}     &
        0.80 ± 0.4 &    \textbf{2084.30 ± 749.91} &  0.005 ± 0.002 \\
        \textit{S}  &
        0.84 ± 0.36 &  2104.13 ± 1022.43 &  0.009 ± 0.005  \\
        \bottomrule
        \end{tabular}
        \end{adjustbox}
    \end{subtable}
    \quad
    \begin{subtable}{.47\linewidth}
    \caption{
    Token-based interpretations decrease human annotation accuracy and increase cognitive load, as indicated by the increased response times. In contrast, sentence-based interpretations significantly improve annotation accuracy and lower the annotators cognitive load. \textit{Sentences lead to the best ITR, a more than 2-fold improvement over tokens}.
    }
    \label{table:wiki_summary}
    \centering
    \begin{adjustbox}{max width=\textwidth}
    \begin{tabular}{lll}
    \toprule
    Accuracy &     Time [s] &      ITR [bits/s]   \\
    \midrule
    0.62 ± 0.49 &  2291.83 ± 1287.84 &  0.0014 ± 0.0018  \\
    0.61 ± 0.49 &  3133.75 ± 1645.26 &  0.0007 ± 0.0012  \\
    \textbf{0.65 ± 0.48} &  \textbf{2046.72 ± 1145.81} &  \textbf{0.0018 ± 0.0017} \\
    \bottomrule
    \end{tabular}
    \end{adjustbox}
    \end{subtable}
\end{table}

%% file: text/conclusion.tex
\section{Conclusions, limitations \& future work}
\label{sec:conclusion}

It has been widely recognized that evaluation of interpretability methods is one of the key research challenges in ML \citep{adebayo_debugging_2020}. 
We complement recent advances in the domain of  computer vision ~\cite{chen_robust_2019,alvarez-melis_towards_2018,kapishnikov_xrai_2019}
and take a first step towards addressing the non-robustness of deep text classifiers. We argue that the large number of tokens means that token-based interpretations are prone to a lack of robustness.
Similarly, large number of tokens may cause high variability 
(\S\ref{sec:explanation_variance}), leading to a degradation of trust by the users.
In line with previous results on the impact of interpretation complexity \citep{lage_evaluation_2018}, our results demonstrate that the non-contiguous nature of token-based interpretations may make them more difficult for the  users to digest, as shown by the human surveys. Our results demonstrate that sentence-based interpretations lead to improved annotation accuracy accompanied by a lower cognitive load to process the interpretations. We find that in difficult tasks, this reduced cognitive load can lead to a substantial increase in annotation throughput. These findings suggest that for certain datasets, sentence-based interpretations have the potential to support AI assisted decision making better than the traditional token-based interpretations and thus contribute effectively to more responsible usage of ML technology.

Finally, we note limitations and avenues for improvement. Owing to their widespread usage, our analysis has been limited to Transformer-based  classifiers only. Extension to other sequence- (\eg, LSTMs/GRUs) and non-sequence-based models (\eg, n-gram models) is an interesting future research direction. An extension to sentence-based interpretability is possible for any (model agnostic) perturbation-based method (\eg, SHAP~\cite{lundberg_unified_2017}, LIME~\cite{ribeiro_why_2016}), however, sentences support only a limited number of gradient-based methods. Specifically, as discussed in \S\ref{sec:metafeature-interp}, only methods for which the feature attributions sum to the output class score---\eg, Integrated Gradients~\cite{sundararajan_axiomatic_2017} and Layerwise Relevance Propagation~\cite{montavon_layer-wise_2019}---are supported. Extending gradient-based methods to accommodate sentence-based interpretability is an important future direction.
Results in \S\ref{sec:human_intelligibility_results} showed that interpretations (token- or sentence-based) do not always improve the annotation performance of humans. We hypothesized that this may be related to the ``difficulty'' level of the task. Further investigations to map dataset characteristics to the kind of interpretations that are more suitable for humans are also worth pursuing.
In this paper, we only focused on sentences as units of meta-token interpretations. We plan to explore extensions to other units such as phrases and paragraphs.
Finally, our work also highlighted the tension between automatic (infidelity) and human subject based evaluation metrics of interpretability. Resolving these tensions and developing semi-automatic and scalable evaluation metrics is another potential direction for research.

%% file: text/appendix.tex
\appendix

\section{Datasets}
\label{app:datasets}

We describe each of the datasets in detail here:

\textbf{IMDB}: The movie review data made publicly available by~\cite{maas_learning_2011}.%
\footnote{\url{https://ai.stanford.edu/~amaas/data/sentiment/}}
The prediction task is to assess whether a given movie review reflects a positive or negative sentiment. The binary sentiment labels were obtained by~\cite{maas_learning_2011} by binning the IMDB review scores that range from 1 to 10. The scores in the range 1-4 are labeled as negative whereas the scores from 7-10 are labeled as positive.

\textbf{Medical}: Medical text data from Kaggle (licensed under CC0: Public Domain).\footnote{\url{https://www.kaggle.com/chaitanyakck/medical-text}} 
The task is to predict the condition of a patient from a medical abstract. The conditions are described by 5 classes: e.g.,  digestive system diseases, cardiovascular diseases, etc.

\textbf{Wiki}: The Wikipedia article data from Kaggle (licensed under CC BY-SA 3.0).\footnote{\url{https://www.kaggle.com/urbanbricks/wikipedia-promotional-articles}} 
The task is to predict whether a Wikipedia article is written with a promotional tone or neutral tone. Examples of `promotional' articles include advertisements, resume-like articles, etc. For more details, we point the reader to the corresponding page on Wikipedia.\footnote{\url{https://en.wikipedia.org/w/index.php?title=Category:Articles_with_a_promotional_tone}} We note that the articles marked promotional tend to be significantly smaller in length than the neutral articles. In order to ensure that the prediction models do not simply use the text length as an indicator or the label, we remove all inputs that have fewer than 500 words.

Table~\ref{table:data_details} shows the statistics for the datasets used in the experiments.
\input{tables/datasets}

\section{Reproducibility} \label{app:reproducability}

\subsection{Software}

We use the HuggingFace~\citep{wolf_transformers_2020} and PyTorch~\citep{paszke_pytorch_2019} libraries for model training. Feature attributions are generated using the Captum library at version 0.4.0~\citep{kokhlikyan_captum_2020}. Input texts were split into sentences using the spaCy library~\cite{honnibal_spacy_2020}. All the libraries are available under open-source licenses at GitHub.

\subsection{Training details}

For training the models, we follow a similar strategy as \cite{zafar_lack_2021}. 

Specifically, we start from pretrained encoders and add a FC-layer of 512 units (with ReLU units) followed by a final classification layer of $C$ units where $C$ is the number of classes. The encoder embeddings are pooled using average pooling before being fed into the FC layer.
As for training, following~\cite{zafar_lack_2021}, we use the AdamW optimizer recently proposed by~\cite{loshchilov_decoupled_2019}. 
Training is run for a maximum of  $25$ epochs. We used the following patience-based early stopping strategy: if the accuracy on the validation set does not increase for $5$ consecutive epochs, the training is stopped.

We divide each dataset into a $80\%-20\%$ train-test split. Furthermore,  $10\%$ of the train set is set aside as a validation set and is used for hyperparameter tuning. The training pipeline consists of two hyperparameters: learning rate which is selected from $\{10^{-2}, 10^{-3}, 10^{-4}, 10^{-5}\}$; and the number of last layers of encoder that should be fine-tuned~\cite{sun_how_2019}, which is selected from $\{0,2\}$.

\section{Model training results}
\label{app:training_results}

Table~\ref{table:acc_models} shows the test set accuracy of the trained (\firstinit) and untrained (\randinit) models.

With the trained model \firstinit, for a given dataset, all the Transformer encoders lead to similar classification accuracy.
For both binary classification datasets (IMDB and Wiki), the accuracy is more than $90\%$ in all cases. For the five-class Medical data, the accuracy is $60\%$.
With the untrained model, as expected, the accuracy is quite low for all the datasets/encoder combinations. One exception to this trend is the Wiki data with \distilroberta model where the test accuracy with \randinit is almost as high as with the trained model \firstinit. However, initializing the \randinit model with other random seeds results in much lower accuracy values---the accuracy values with 5 different random seeds are $0.06$, $0.28$, $0.49$, $0.29$ and $0.71$ (as shown in Table~\ref{table:data_details}, the majority / minority class split in the data is $71\%$ / $29\%$).

Table~\ref{table:common_preds} shows the fraction of common predictions between different initializations.
For both IMDB and Wiki data, the fraction of common predictions between the two trained models is almost $100\%$ meaning that the two models are indeed almost functionally equivalent~\citep{sundararajan_axiomatic_2017,zafar_lack_2021}. For the Medical data, the fraction of common predictions is somewhat lower.

\input{tables/accuracy}
\input{tables/frac_common_predictions}

\section{Robustness and variability with indirect computation of sentence attributions} \label{app:sentence_indirect}

Here, we report the analogue of results from Section~\ref{sec:robustness_tests} when the sentence-level attribution scores are derived via the indirect method in Section~\ref{sec:metafeature-interp} by summing the contributions of the individual tokens.

\input{tables/shap_indirect_jaccard_difference}
Table~\ref{table:shap_indirect_jac_diff} shows the results of the two randomization tests for SHAP. In a manner similar to that in Section~\ref{sec:robustness_tests}, the sentence-based interpretations perform better in both the randomization tests.

\input{tables/IG_indirect_jaccard_difference}
Table~\ref{table:ig_indirect_jac_diff} shows the results of the two randomization tests for IG. While the results for the \AbbrvTestdiffinit are similar to those for SHAP (sentence-based interpretations are more robust), the results for the \AbbrvTestuntrained show an opposite trend, the token-based interpretations are more robust.

\input{tables/shap_indirect_multirun_jaccard}
Table~\ref{table:shap_indirect_jac_multiruns} shows the median difference in overlap metric between sentences and tokens. As opposed to direct computation in \S\ref{sec:explanation_variance}, we notice a higher variability with sentences as compared to tokens.

\section{Human intelligibility surveys} \label{app:surveys}

\subsection{Data pre-processing}
\label{app:survey_data_processing}

We applied two pre-processing steps to the texts before the annotations tasks:

\begin{enumerate}
    \item For token-level highlighting, we noticed that the individual features for the Transformer models (that is, the tokens) can be at an even smaller granularity than words. This happens due to the sub-word tokenization employed by these models to reduce the size of the embedding matrix. This subword tokenization results in cases where the words may get split in the following manner: `ailments' $\to$ `ai' + `\#\#lm' + `\#\#ents'~\citep{devlin_bert_2019}. In order to avoid any additional complexity arising from this subword tokenization, we merge any sub-word tokens into a whole word and use the average of the subword token attributions as the attribution of the merged word.
    \item We notice that highlighting sentences may result in cases where more than 10\% of the text gets highlighted (due to the sentence being longer than 10\% of the text). For a fair comparison, we truncate such overflowing sentences such that no more than 10\% of the text is highlighted.
\end{enumerate}

\subsection{Annotator instructions} \label{app:annotator_instructions}

\begin{figure}
     \centering
     \begin{subfigure}[b]{0.90\textwidth}
         \centering
        \includegraphics[width=\textwidth]{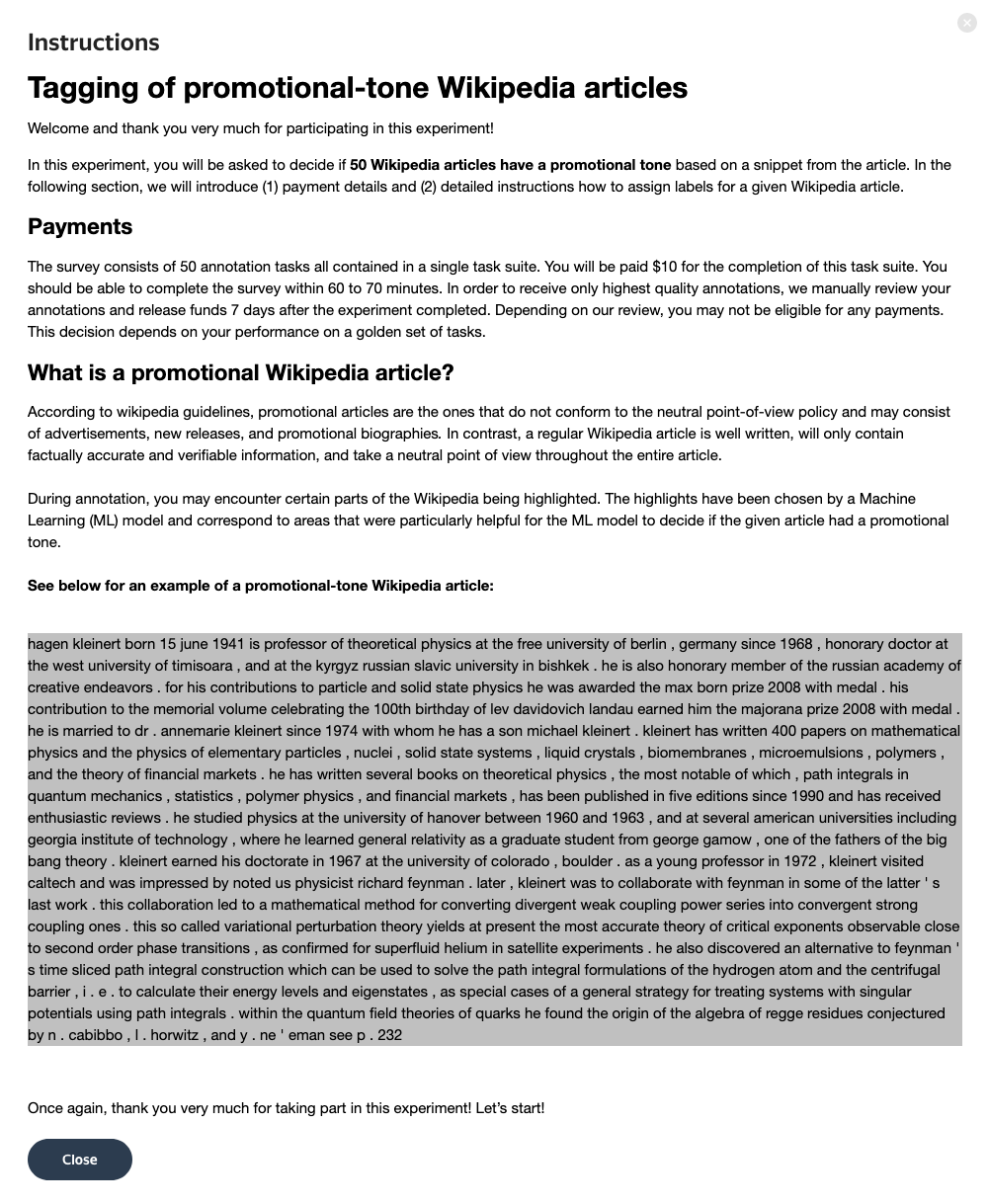}
         \caption{Landing page for the survey with the Wiki data.}
         \label{fig:landing_wiki}
     \end{subfigure}
     \\
     \begin{subfigure}[b]{0.90\textwidth}
         \centering
        \includegraphics[width=\textwidth]{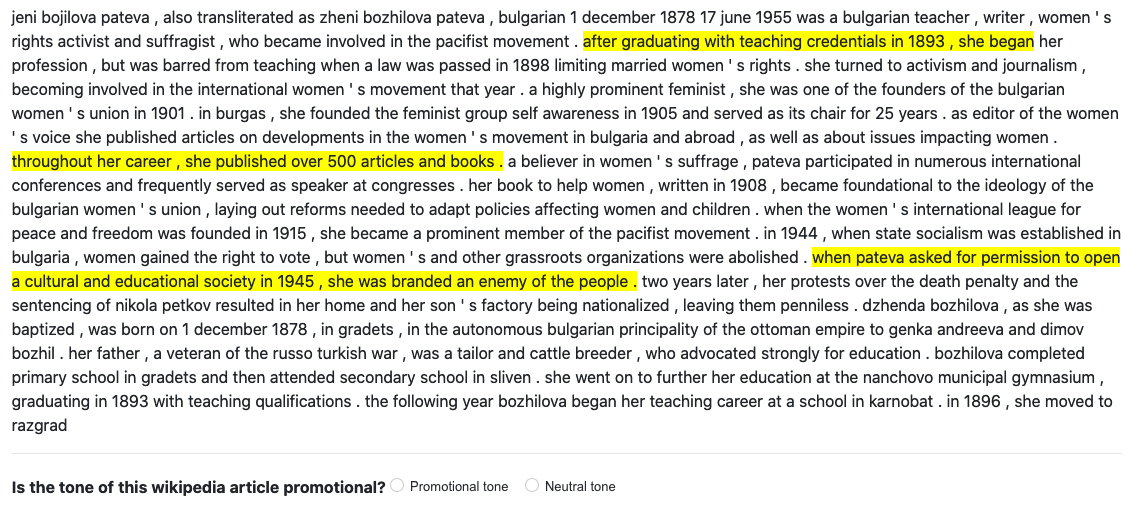}
         \caption{Example of a highlighted text for the Wiki data.}
         \label{fig:sentence_highlighted_wiki}
     \end{subfigure}
        \caption{Examples of the landing page and the highlighted text from the Wikipedia human surveys.}
        \label{fig:wiki_survey_examples}
\end{figure}

\begin{figure}
     \centering
     \begin{subfigure}[b]{0.95\textwidth}
         \centering
        \includegraphics[width=\textwidth]{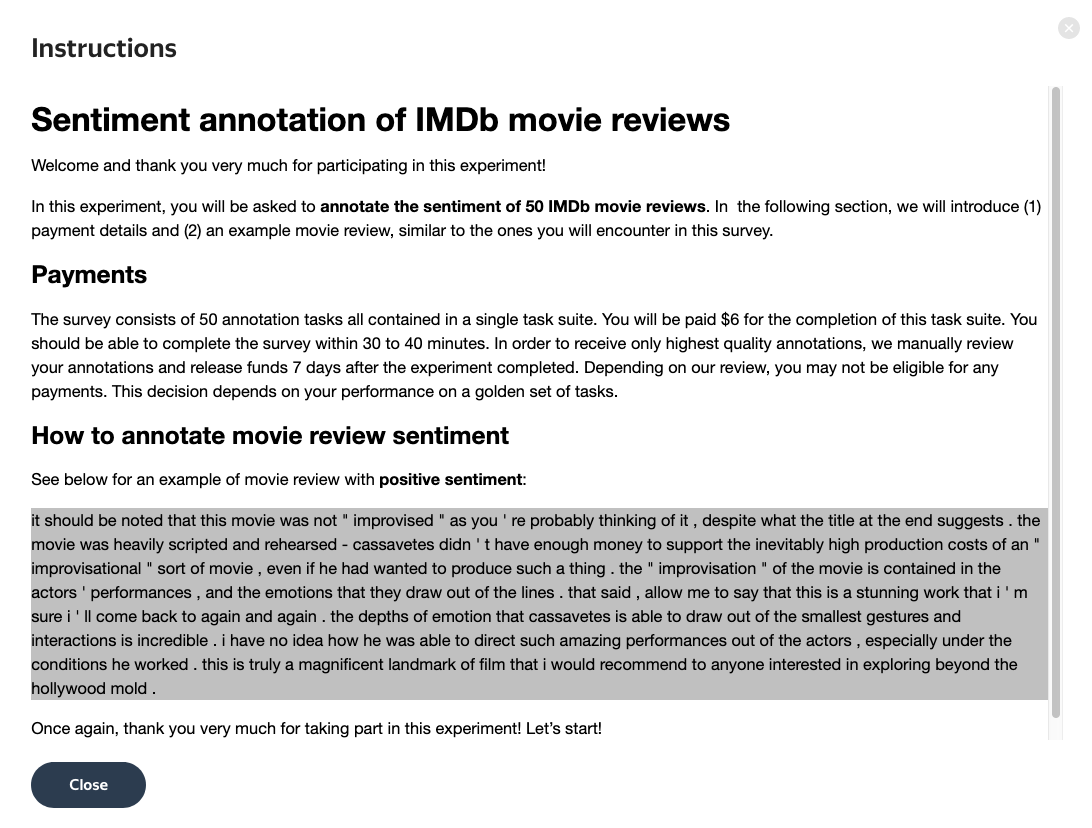}
         \caption{Landing page for the survey with the IMDB data.}
         \label{fig:landing_imdb}
     \end{subfigure}
     \\
     \begin{subfigure}[b]{0.95\textwidth}
         \centering
        \includegraphics[width=\textwidth]{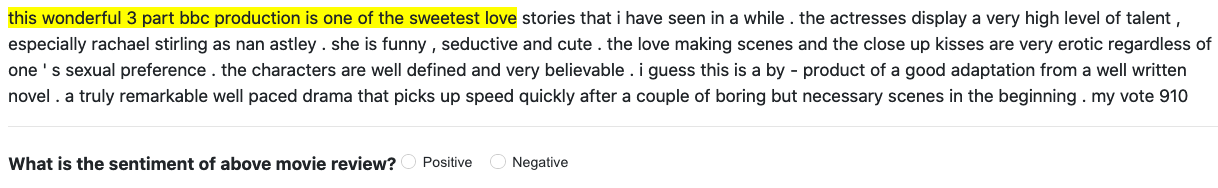}
         \caption{Example of a highlighted text for the IMDB data.}
         \label{fig:sentence_highlighted_imdb}
     \end{subfigure}
        \caption{Examples of the landing page and the highlighted text from the IMDB human surveys.}
        \label{fig:imdb_survey_examples}
\end{figure}

Figure~\ref{fig:wiki_survey_examples} and Figure~\ref{fig:imdb_survey_examples} shows the examples of a landing page and a highlighted text on the Wiki and IMDB datasets respectively.

\subsection{Payment details} \label{app:toloka_payments}

We ran six user surveys in total, one for IMDB and another for the Wiki dataset, and three conditions for each data (control, token highlighted, sentence highlighted).  Users were randomly assigned to each condition. Each survey consisted of 50 individual questions (one for each input text).  
In the end, for the IMDB data, we gathered 750, 750 and 650 annotations for the control, sentence and token conditions, respectively. For Wiki, the numbers were 600, 750 and 600.

To be eligible for payments, annotators had to complete all 50 tasks. For both surveys, we calibrated payments for an hourly wage of \$11. As payments were done for completed surveys of 50 tasks, we estimated single task completion times of about 30 seconds and 60 seconds for IMDB and Wiki datasets respectively. Users received \$6 to complete the IMDb survey and \$10 for the Wikipedia survey. In hindsight, users did complete tasks faster than what we estimated and therefore the initially set hourly wage of \$11 was exceeded.

\subsection{Implementation of Information Transfer Rate (ITR)}

Following from \S\ref{sec:human_intel_metrics}, let $y \in \{0,1\}^N$ denote the vector of true labels of the texts that a human annotator was asked to annotate, and let $y_h \in \{0,1\}^N$ denote the vector of annotations provided by the human, where $N$ is the number of questions in the survey. Then the ITR is computed as~\cite{schmidt_quantifying_2019}: $ITR = \frac{I(y_h, y)}{t}$, where $t$ is the average response time for each question, and $I$ is the mutual information.

\input{tables/fidelity_word}

\input{tables/fidelity_sentence}

For implementing $I$, we use the Mutual-Information Score function from \texttt{scikit-learn}.%
\footnote{\url{https://scikit-learn.org/stable/modules/generated/sklearn.metrics.mutual_info_score.html}}
Specifically:
$I(y_h,y)=\sum_{i \in \{0,1\}} \sum_{j \in \{0,1\}} \frac{|y_h^{(i)} \cap y^{(j)}|}{N} \log \frac{N |y_h^{(i)} \cap y^{(j)}|}{|y_h^{(i)}||y^{(j)}|}$,
where $y_h^{(i)}$ denotes the indices of $y_h$ with value $i$.

\section{(Mis)alignment between human and computational metrics.}
\label{app:infidelity}

Here, we compare the human annotation performance with a closely related measure of \textit{infidelity}. The infidelity measure is often used to automatically (without human assistance) measure how important the top-ranked features (by an attribution method) are to the model output~\cite{zafar_lack_2021,atanasova_diagnostic_2020,fong_understanding_2019,samek_evaluating_2017,lundberg_unified_2017,arras_explaining_2016}. 
The metric is computed as follows:
Given the feature attributions (\eg, token- or sentence-based), remove the features from the input iteratively in the order of decreasing importance until the model prediction changes. Then, infidelity is defined as the percentage of text that need to be dropped for the prediction to change. Thus, while ITR describes how well the feature attribution ranking aligns with human perception, infidelity describes how well the attributions align with the model itself.

The infidelity metric is used in a number of domains (tabular, image, text) and appears in many closely related variations, all aiming to measure the change in the model prediction upon dropping important features~\cite{zafar_lack_2021,atanasova_diagnostic_2020,fong_understanding_2019,samek_evaluating_2017,lundberg_unified_2017,arras_explaining_2016}. We use the text-classification variant used in~\cite{zafar_lack_2021,arras_explaining_2016}.
We count the percentage of text that needs to be dropped in terms of tokens in the input (for both token as well as sentence-based interpretations).

\begin{figure}[ht]
     \centering
     \begin{subfigure}[b]{0.45\textwidth}
         \centering
        \includegraphics[width=\textwidth]{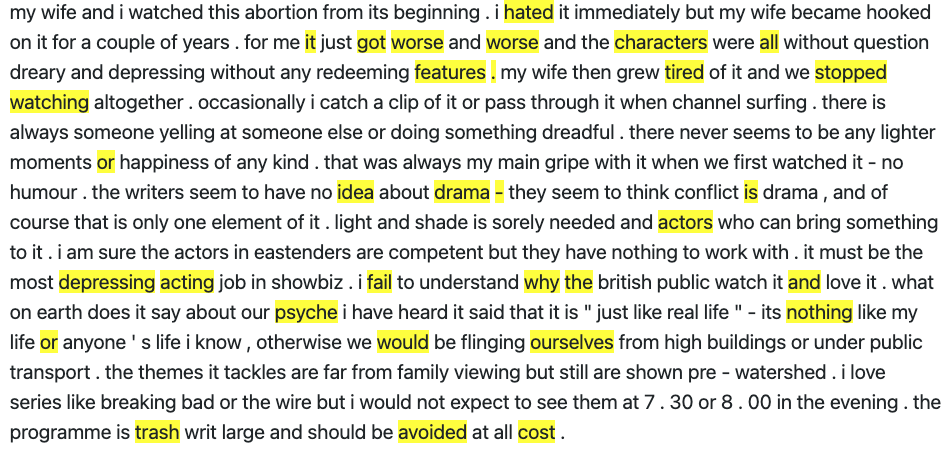}
         \caption{Tokens.}
         \label{fig:imdb_word_highlighted}
     \end{subfigure}
     \hfill
     \begin{subfigure}[b]{0.45\textwidth}
         \centering
        \includegraphics[width=\textwidth]{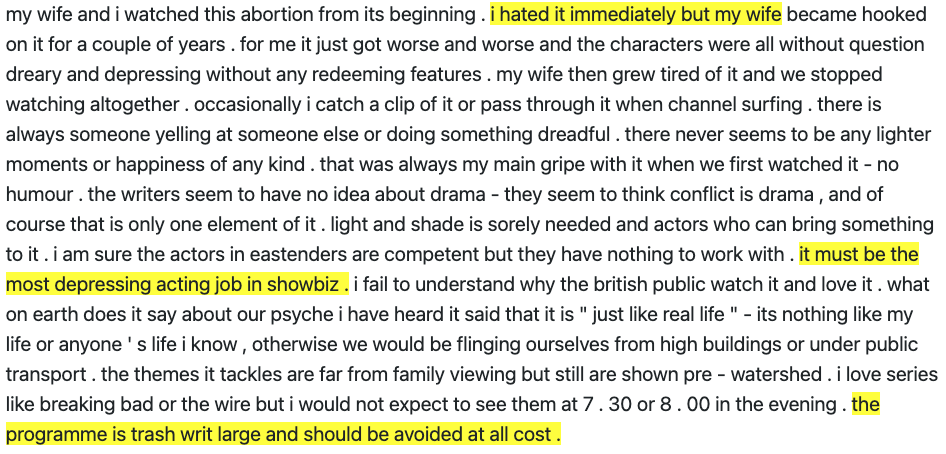}
         \caption{Sentences.}
         \label{fig:imdb_sentence_highlighted}
     \end{subfigure}
        \caption{Highlighting 10\% of the text with token and sentence-based interpretations.}
        \label{fig:imdb_highlighted}
\end{figure}

\paragraph{Results}
Table~\ref{table:fidelity_best_word} shows the infidelity of the token-based interpretations for the trained model (\firstinit). Table~\ref{table:fidelity_best_sentence} shows the infidelity for sentence-based interpretations. 

We notice that the infidelity of the token-based interpretations with IMDB data is 15\% whereas that of sentence-based interpretations is 61\% (a lower infidelity is better as fewer important features need to be dropped for the prediction to switch).  The token- and sentence-based infidelity for the Wiki data are 38\% and 57\% respectively. In other words, quite surprisingly, while the sentence-based interpretations lead to better human annotation performance, they do not necessarily perform better according to a related computational metric used by several studies. We see similar insights for other datasets and models. 

We show one possible reason for this via an example in Figure~\ref{fig:imdb_highlighted}. 
The figure shows the same instance with top-10\% features highlighted with both token and sentence-based interpretations. When highlighting sentences-based interpretations, we truncate the sentence if it overflows beyond 10\% of the tokens in the text (Appendix~\ref{app:survey_data_processing}).
While the sentence-based highlights are naturally quite compact, the token-based interpretations show that important tokens are spread all over the document.
Such redundancies could require nearly all these sentences to be dropped for the model prediction to change.

%% file: tables/datasets.tex
\begin{table}[h]
\caption{Detailed statistics of the datasets used in the experiments. The columns Words and Sentences show the average $\pm$ standard deviation across the data. Prev. Most shows the prevalence (in percentage) of the most prevalent class in the dataset.}
\centering
\label{table:data_details}
\begin{tabular}{lcccccc}\\\toprule  
Dataset & Samples & Classes & Prev. Most & Prev. Least & Words & Sentences  \\\midrule
IMDB & 
50,000  &   2 &     50 &    50 &    232$\pm$171 &    10$\pm$7 
\\  
Medical & 
14,438 &    5 &     33 &    10 &   184$\pm$80 &     8$\pm$3
\\ 
Wiki &
41,460 &    2 &    71 &     29 &  2,603$\pm$1,618 & 117$\pm$74\\  \bottomrule
\end{tabular}
\end{table}

%% file: tables/accuracy.tex
\begin{table}[t]
    \caption{
    Test accuracy with best model (\firstinit) and the untrained model (\randinit). 
    }
    \label{table:acc_models}
    \begin{subtable}{.5\linewidth}
      \caption{\firstinit}
      \centering
        \begin{tabular}{lrrrr}
        \toprule
        {} &  \bert &  \roberta &  \distilbert &  \distilroberta \\
        \midrule
        \textbf{IMDB    } &   0.92 &      0.95 &         0.93 &            0.94 \\
        \textbf{Medical } &   0.60 &      0.64 &         0.63 &            0.65 \\
        \textbf{Wiki    } &   0.94 &      0.95 &         0.95 &            0.96 \\
        \bottomrule
        \end{tabular}
        \label{table:acc_init1}
    \end{subtable}%
    \quad
    \begin{subtable}{0.5\linewidth}
      \centering
    \caption{\randinit}
    \begin{tabular}{lrrrr}
        \toprule
        {} &  \bert &  \roberta &  \distilbert &  \distilroberta \\
        \midrule
        \textbf{IMDB    } &   0.27 &      0.51 &         0.20 &            0.46 \\
        \textbf{Medical } &   0.21 &      0.28 &         0.04 &            0.07 \\
        \textbf{Wiki    } &   0.65 &      0.05 &         0.78 &            0.94 \\
        \bottomrule
        \end{tabular}
        \label{table:acc_untrained}
    \end{subtable} 
\end{table}

%% file: tables/frac_common_predictions.tex
\begin{table}[t]
    \caption{Percentage of common predictions between different initializations.}
    \label{table:common_preds}
    \begin{subtable}{.5\linewidth}
      \caption{\firstinit vs. \secondinit}
      \centering
        \begin{tabular}{lrrrr}
        \toprule
        {} &  \bert &  \roberta &  \distilbert &  \distilroberta \\
        \midrule
        \textbf{IMDB    } &     98 &        97 &           97 &              98 \\
        \textbf{Medical } &     88 &        80 &           82 &              91 \\
        \textbf{Wiki    } &     98 &        97 &           98 &              96 \\
        \bottomrule
        \end{tabular}
        \label{table:common_preds_init1_init2}
    \end{subtable}%
    \quad
    \begin{subtable}{.5\linewidth}
      \centering
    \caption{\firstinit vs. \randinit}
    \begin{tabular}{lrrrr}
        \toprule
        {} &  \bert &  \roberta &  \distilbert &  \distilroberta \\
        \midrule
        \textbf{IMDB    } &     26 &        50 &           18 &              45 \\
        \textbf{Medical } &     25 &        35 &            2 &               5 \\
        \textbf{Wiki    } &     68 &         5 &           81 &              97 \\
        \bottomrule
        \end{tabular}
        \label{table:common_preds_init1_untrained}
    \end{subtable} 
\end{table}

%% file: tables/shap_indirect_jaccard_difference.tex
\begin{table}[ht]
    \caption{[SHAP indirect computation for sentence-based interpretations] Positive (median) values of $J_s - J_t$ (\S\ref{sec:stability_metric}) for \AbbrvTestdiffinit in 8/12, and negative values for \AbbrvTestuntrained in 10/12 cases mean that sentence-based interpretations are more robust w.r.t. both tests.}
    \label{table:shap_indirect_jac_diff}
    \begin{subtable}{.5\linewidth}
      \caption{\AbbrvTestdiffinit: Median difference in Jaccard@25\% of sentences and tokens when comparing over \firstinit and \secondinit. Positive values mean sentences are more robust.}
      \centering
        \begin{tabular}{lrrrr}
        \toprule
        {} &  \bert &  \roberta &  \distilbert &  \distilroberta \\
        \midrule
        \textbf{IMDB   } &     17 &        -3 &           10 &              17 \\
        \textbf{Medical} &     14 &        -5 &            9 &              15 \\
        \textbf{Wiki   } &      5 &       -12 &            5 &              -3 \\
        \bottomrule
        \end{tabular}
        \label{table:shap_indirect_jac_diff_init1_init2}
    \end{subtable}%
    \quad
    \begin{subtable}{.5\linewidth}
      \centering
    \caption{\AbbrvTestuntrained: Median difference in Jaccard@25\% of sentences and tokens when comparing over \firstinit and \randinit. Negative values mean sentences are more robust.}
    \begin{tabular}{lrrrr}
        \toprule
        {} &  \bert &  \roberta &  \distilbert &  \distilroberta \\
        \midrule
        \textbf{IMDB   } &     -8 &        13 &           -5 &             -12 \\
        \textbf{Medical} &     -5 &        -2 &            4 &             -10 \\
        \textbf{Wiki   } &     -9 &       -12 &          -10 &              -3 \\
        \bottomrule
        \end{tabular}
        \label{table:shap_indirect_jac_diff_init1_untrained}
    \end{subtable} 
\end{table}

%% file: tables/IG_indirect_jaccard_difference.tex
\begin{table}[t]
    \caption{[IG indirect computation for sentence-based interpretations] Positive (median) values of $J_s - J_t$ (\S\ref{sec:stability_metric}) for \AbbrvTestdiffinit in 8/12 cases means that sentence-based interpretations are more robust. However, as opposed to SHAP, in 10/12 cases, \AbbrvTestuntrained results in positive values, showing that the token-based interpretations are more robust.}

    \label{table:ig_indirect_jac_diff}
    \begin{subtable}{.5\linewidth}
      \caption{\AbbrvTestdiffinit: Median difference in Jaccard@25\% of sentences and tokens when comparing over \firstinit and \secondinit. Positive values mean sentences are more robust.}
      \centering
        \begin{tabular}{lrrrr}
        \toprule
        {} &  \bert &  \roberta &  \distilbert &  \distilroberta \\
        \midrule
        \textbf{IMDB   } &     16 &        -2 &           13 &              23 \\
        \textbf{Medical} &     23 &         3 &           23 &              21 \\
        \textbf{Wiki   } &     -2 &        -9 &           -7 &               0 \\
        \bottomrule
        \end{tabular}
        \label{table:ig_indirect_jac_diff_init1_init2}
    \end{subtable}%
    \quad
    \begin{subtable}{.5\linewidth}
      \centering
    \caption{\AbbrvTestuntrained: Median difference in Jaccard@25\% of sentences and tokens when comparing over \firstinit and \randinit. Negative values mean sentences are more robust.}
    \begin{tabular}{lrrrr}
        \toprule
        {} &  \bert &  \roberta &  \distilbert &  \distilroberta \\
        \midrule
        \textbf{IMDB   } &      8 &        16 &            9 &              10 \\
        \textbf{Medical} &      0 &        21 &           39 &              16 \\
        \textbf{Wiki   } &     13 &        -2 &           -4 &               4 \\
        \bottomrule
        \end{tabular}
        \label{table:ig_indirect_jac_diff_init1_untrained}
    \end{subtable} 
\end{table}

%% file: tables/shap_indirect_multirun_jaccard.tex
\begin{table}[t]
    \caption{[SHAP indirect computation for sentence-based interpretations]
    Median difference between overlap (\S\ref{sec:stability_metric}) of sentences and tokens when computing the feature attributions repeatedly 10 times. A positive value means that sentences-based attributions are more stable.}
    \label{table:shap_indirect_jac_multiruns}
    \begin{subtable}{.5\linewidth}
      \caption{Trained model (\firstinit)}
      \centering
        \begin{tabular}{lrrrr}
        \toprule
        {} &  \bert &  \roberta &  \distilbert &  \distilroberta \\
        \midrule
        \textbf{IMDB   } &                 -5 &             0 &                       -2 &                  -3 \\
        \textbf{Medical} &                 -2 &            -1 &                       13 &                  13 \\
        \textbf{Wiki   } &                -17 &            -6 &                      -11 &                   8 \\
        \bottomrule
        \end{tabular}
        \label{table:shap_indirect_jac_multiruns_trained}
    \end{subtable}%
    \quad
    \begin{subtable}{.5\linewidth}
      \centering
    \caption{Untrained model (\randinit)}
    \begin{tabular}{lrrrr}
        \toprule
        {} &  \bert &  \roberta &  \distilbert &  \distilroberta \\
        \midrule
        \textbf{IMDB   } &                  2 &            -1 &                        2 &                  -1 \\
        \textbf{Medical} &                 -2 &            -3 &                       -4 &                  -4 \\
        \textbf{Wiki   } &                -10 &            -7 &                      -11 &                   0 \\
        \bottomrule
        \end{tabular}
        \label{table:shap_indirect_jac_untrained}
    \end{subtable} 
\end{table}

%% file: tables/fidelity_word.tex
\begin{table*}[t]
\caption{Mean infidelity of different interpretability methods at \textit{token-level}. Lower values are better.}
\centering
\small
\begin{tabular}{l|cccc|cccc|cccc}
\toprule
& 
\multicolumn{4}{|c}{\bf IMDB} &
\multicolumn{4}{|c}{\bf Medical} &
\multicolumn{4}{|c}{\bf Wiki} \\
{} 
&  \bert &  \roberta &  \distilbert &  \distilroberta 
&  \bert &  \roberta &  \distilbert &  \distilroberta 
&  \bert &  \roberta &  \distilbert &  \distilroberta 
\\
\midrule
\textbf{SHAP} 
&     15 &        28 &           22 &              18 
&     25 &        23 &           22 &              17 
&     38 &        39 &           37 &              31 
\\
\textbf{IG} 
&     39 &        44 &           45 &              47 
&     26 &        22 &           25 &              18 
&     40 &        43 &           38 &              46 
\\
\bottomrule
\end{tabular}
\label{table:fidelity_best_word}
\end{table*}

%% file: tables/fidelity_sentence.tex
\begin{table*}[t]
\caption{Mean infidelity of different interpretability methods at \textit{sentence-level}. Suffix ``-i'' denotes the indirect computation whereas ``-d'' denotes the direct computation (\S\ref{sec:generating_metatoken}). Lower values are better.}
\centering
\small
\begin{tabular}{l|cccc|cccc|cccc}
\toprule
& 
\multicolumn{4}{|c}{\bf IMDB} &
\multicolumn{4}{|c}{\bf Medical} &
\multicolumn{4}{|c}{\bf Wiki} \\
{} 
&  \bert &  \roberta &  \distilbert &  \distilroberta 
&  \bert &  \roberta &  \distilbert &  \distilroberta 
&  \bert &  \roberta &  \distilbert &  \distilroberta 
\\
\midrule
\textbf{SHAP-d} 
&     61 &        63 &           65 &              63 
&     55 &        59 &           58 &              59 
&     57 &        50 &           58 &              52 \\
\textbf{SHAP-i} 
&     68 &        72 &           70 &              71 
&     62 &        64 &           63 &              63 
&     69 &        67 &           72 &              65 \\
\textbf{IG-i} 
&     78 &        79 &           76 &              78 
&     60 &        65 &           62 &              67 
&     67 &        67 &           74 &              70
\\
\bottomrule
\end{tabular}
\label{table:fidelity_best_sentence}
\end{table*}

%% file: main.bbl
\begin{thebibliography}{10}

\bibitem{abid_persistent_2021}
Abubakar Abid, Maheen Farooqi, and James Zou.
\newblock Persistent {Anti}-{Muslim} {Bias} in {Large} {Language} {Models}.
\newblock {\em arXiv:2101.05783 [cs]}, January 2021.
\newblock arXiv: 2101.05783.

\bibitem{adebayo_sanity_2018}
Julius Adebayo, Justin Gilmer, Michael Muelly, Ian Goodfellow, Moritz Hardt,
  and Been Kim.
\newblock Sanity {Checks} for {Saliency} {Maps}.
\newblock In {\em Proceedings of the 32nd {International} {Conference} on
  {Neural} {Information} {Processing} {Systems}}, {NIPS}'18, pages 9525--9536,
  Red Hook, NY, USA, December 2018. Curran Associates Inc.

\bibitem{adebayo_debugging_2020}
Julius Adebayo, Michael Muelly, Ilaria Liccardi, and Been Kim.
\newblock Debugging {Tests} for {Model} {Explanations}.
\newblock In {\em Proceedings of the 34th {International} {Conference} on
  {Neural} {Information} {Processing} {Systems}}, 2020.

\bibitem{alvarez-melis_towards_2018}
David Alvarez-Melis and Tommi~S. Jaakkola.
\newblock Towards {Robust} {Interpretability} with {Self}-explaining {Neural}
  {Networks}.
\newblock In {\em Proceedings of the 32nd {International} {Conference} on
  {Neural} {Information} {Processing} {Systems}}, {NIPS}'18, pages 7786--7795,
  Montréal, Canada, December 2018. Curran Associates Inc.

\bibitem{arras_explaining_2016}
Leila Arras, Franziska Horn, Grégoire Montavon, Klaus-Robert Müller, and
  Wojciech Samek.
\newblock Explaining {Predictions} of {Non}-{Linear} {Classifiers} in {NLP}.
\newblock In {\em Proceedings of the 1st {Workshop} on {Representation}
  {Learning} for {NLP}}, pages 1--7, June 2016.

\bibitem{arras_evaluating_2019}
Leila Arras, Ahmed Osman, Klaus-Robert Müller, and Wojciech Samek.
\newblock Evaluating {Recurrent} {Neural} {Network} {Explanations}.
\newblock {\em arXiv:1904.11829 [cs, stat]}, June 2019.

\bibitem{atanasova_diagnostic_2020}
Pepa Atanasova, Jakob~Grue Simonsen, Christina Lioma, and Isabelle Augenstein.
\newblock A {Diagnostic} {Study} of {Explainability} {Techniques} for {Text}
  {Classification}.
\newblock In {\em Proceedings of the 2020 {Conference} on {Empirical} {Methods}
  in {Natural} {Language} {Processing} ({EMNLP})}, pages 3256--3274, Online,
  November 2020. Association for Computational Linguistics.

\bibitem{beltagy_scibert_2019}
Iz~Beltagy, Kyle Lo, and Arman Cohan.
\newblock {SciBERT}: {A} {Pretrained} {Language} {Model} for {Scientific}
  {Text}.
\newblock {\em arXiv:1903.10676 [cs]}, September 2019.

\bibitem{bender_dangers_2021}
Emily~M. Bender, Timnit Gebru, Angelina McMillan-Major, and Shmargaret
  Shmitchell.
\newblock On the {Dangers} of {Stochastic} {Parrots}: {Can} {Language} {Models}
  {Be} {Too} {Big}? \&\#x1f99c;.
\newblock In {\em Proceedings of the 2021 {ACM} {Conference} on {Fairness},
  {Accountability}, and {Transparency}}, {FAccT} '21, pages 610--623, Virtual
  Event, Canada, March 2021. Association for Computing Machinery.

\bibitem{bhatt_explainable_2020}
Umang Bhatt, Alice Xiang, Shubham Sharma, Adrian Weller, Ankur Taly, Yunhan
  Jia, Joydeep Ghosh, Ruchir Puri, José M.~F. Moura, and Peter Eckersley.
\newblock Explainable {Machine} {Learning} in {Deployment}.
\newblock In {\em Proceedings of the 2020 {Conference} on {Fairness},
  {Accountability}, and {Transparency}}, {FAT}* '20, pages 648--657, Barcelona,
  Spain, January 2020. Association for Computing Machinery.

\bibitem{breiman_random_2001}
Leo Breiman.
\newblock Random {Forests}.
\newblock {\em Machine Learning}, 45(1):5--32, October 2001.

\bibitem{brown_language_2020}
Tom Brown, Benjamin Mann, Nick Ryder, Melanie Subbiah, Jared~D Kaplan, Prafulla
  Dhariwal, Arvind Neelakantan, Pranav Shyam, Girish Sastry, Amanda Askell,
  Sandhini Agarwal, Ariel Herbert-Voss, Gretchen Krueger, Tom Henighan, Rewon
  Child, Aditya Ramesh, Daniel Ziegler, Jeffrey Wu, Clemens Winter, Chris
  Hesse, Mark Chen, Eric Sigler, Mateusz Litwin, Scott Gray, Benjamin Chess,
  Jack Clark, Christopher Berner, Sam McCandlish, Alec Radford, Ilya Sutskever,
  and Dario Amodei.
\newblock Language {Models} are {Few}-{Shot} {Learners}.
\newblock In H.~Larochelle, M.~Ranzato, R.~Hadsell, M.~F. Balcan, and H.~Lin,
  editors, {\em Advances in {Neural} {Information} {Processing} {Systems}},
  volume~33, pages 1877--1901. Curran Associates, Inc., 2020.

\bibitem{chen_robust_2019}
Jiefeng Chen, Xi~Wu, Vaibhav Rastogi, Yingyu Liang, and Somesh Jha.
\newblock Robust {Attribution} {Regularization}.
\newblock In {\em Advances in {Neural} {Information} {Processing} {Systems}},
  volume~32, 2019.

\bibitem{covert_improving_2021}
Ian Covert and Su-In Lee.
\newblock Improving {KernelSHAP}: {Practical} {Shapley} {Value} {Estimation}
  via {Linear} {Regression}.
\newblock {\em arXiv:2012.01536 [cs, stat]}, April 2021.
\newblock arXiv: 2012.01536.

\bibitem{devlin_bert_2019}
Jacob Devlin, Ming-Wei Chang, Kenton Lee, and Kristina Toutanova.
\newblock {BERT}: {Pre}-training of {Deep} {Bidirectional} {Transformers} for
  {Language} {Understanding}.
\newblock In {\em Proceedings of the 2019 {Conference} of the {North}
  \{{A}\}merican {Chapter} of the {Association} for {Computational}
  {Linguistics}: {Human} {Language} {Technologies}, {Volume} 1 ({Long} and
  {Short} {Papers})}, May 2019.

\bibitem{deyoung_eraser_2020}
Jay DeYoung, Sarthak Jain, Nazneen~Fatema Rajani, Eric Lehman, Caiming Xiong,
  Richard Socher, and Byron~C. Wallace.
\newblock {ERASER}: {A} {Benchmark} to {Evaluate} {Rationalized} {NLP}
  {Models}.
\newblock In {\em Proceedings of the 58th {Annual} {Meeting} of the
  {Association} for {Computational} {Linguistics}}, pages 4443--4458, Online,
  July 2020. Association for Computational Linguistics.

\bibitem{dhamala_bold_2021}
Jwala Dhamala, Tony Sun, Varun Kumar, Satyapriya Krishna, Yada Pruksachatkun,
  Kai-Wei Chang, and Rahul Gupta.
\newblock {BOLD}: {Dataset} and {Metrics} for {Measuring} {Biases} in
  {Open}-{Ended} {Language} {Generation}.
\newblock In {\em Proceedings of the 2021 {ACM} {Conference} on {Fairness},
  {Accountability}, and {Transparency}}, {FAccT} '21, pages 862--872, Virtual
  Event, Canada, March 2021. Association for Computing Machinery.

\bibitem{doshi-velez_towards_2017}
Finale Doshi-Velez and Been Kim.
\newblock Towards {A} {Rigorous} {Science} of {Interpretable} {Machine}
  {Learning}.
\newblock {\em arXiv:1702.08608 [cs, stat]}, March 2017.
\newblock arXiv: 1702.08608.

\bibitem{fedus_switch_2021}
William Fedus, Barret Zoph, and Noam Shazeer.
\newblock Switch {Transformers}: {Scaling} to {Trillion} {Parameter} {Models}
  with {Simple} and {Efficient} {Sparsity}.
\newblock {\em arXiv:2101.03961 [cs]}, January 2021.
\newblock arXiv: 2101.03961.

\bibitem{feng_pathologies_2018}
Shi Feng, Eric Wallace, Alvin Grissom~II, Mohit Iyyer, Pedro Rodriguez, and
  Jordan Boyd-Graber.
\newblock Pathologies of {Neural} {Models} {Make} {Interpretations}
  {Difficult}.
\newblock In {\em Proceedings of the 2018 {Conference} on {Empirical} {Methods}
  in {Natural} {Language} {Processing}}, August 2018.

\bibitem{fong_understanding_2019}
Ruth Fong, Mandela Patrick, and Andrea Vedaldi.
\newblock Understanding {Deep} {Networks} via {Extremal} {Perturbations} and
  {Smooth} {Masks}.
\newblock In {\em Proceedings of the {IEEE}/{CVF} {International} {Conference}
  on {Computer} {Vision} ({ICCV})}, pages 2950--2958, 2019.

\bibitem{gilpin_explaining_2018}
L.~H. Gilpin, D.~Bau, B.~Z. Yuan, A.~Bajwa, M.~Specter, and L.~Kagal.
\newblock Explaining {Explanations}: {An} {Overview} of {Interpretability} of
  {Machine} {Learning}.
\newblock In {\em 2018 {IEEE} 5th {International} {Conference} on {Data}
  {Science} and {Advanced} {Analytics} ({DSAA})}, pages 80--89, October 2018.

\bibitem{guidotti_survey_2018}
Riccardo Guidotti, Anna Monreale, Salvatore Ruggieri, Franco Turini, Fosca
  Giannotti, and Dino Pedreschi.
\newblock A {Survey} of {Methods} for {Explaining} {Black} {Box} {Models}.
\newblock {\em ACM Computing Surveys}, 51(5):93:1--93:42, August 2018.

\bibitem{hase_evaluating_2020}
Peter Hase and Mohit Bansal.
\newblock Evaluating {Explainable} {AI}: {Which} {Algorithmic} {Explanations}
  {Help} {Users} {Predict} {Model} {Behavior}?
\newblock In {\em Proceedings of the 58th {Annual} {Meeting} of the
  {Association} for {Computational} {Linguistics}}, pages 5540--5552, Online,
  July 2020. Association for Computational Linguistics.

\bibitem{honnibal_spacy_2020}
Matthew Honnibal, Ines Montani, Sofie Van~Landeghem, and Adriane Boyd.
\newblock {spaCy}: {Industrial}-strength {Natural} {Language} {Processing} in
  {Python}, 2020.
\newblock https://spacy.io/.

\bibitem{janzing_causal_2019}
Dominik Janzing, Kailash Budhathoki, Lenon Minorics, and Patrick Blöbaum.
\newblock Causal structure based root cause analysis of outliers.
\newblock {\em arXiv:1912.02724 [cs, math, stat]}, December 2019.

\bibitem{kapishnikov_xrai_2019}
Andrei Kapishnikov, Tolga Bolukbasi, Fernanda Viegas, and Michael Terry.
\newblock {XRAI}: {Better} {Attributions} {Through} {Regions}.
\newblock In {\em 2019 {IEEE}/{CVF} {International} {Conference} on {Computer}
  {Vision} ({ICCV})}, pages 4947--4956, Seoul, Korea (South), October 2019.
  IEEE.

\bibitem{kokhlikyan_captum_2020}
Narine Kokhlikyan, Vivek Miglani, Miguel Martin, Edward Wang, Bilal Alsallakh,
  Jonathan Reynolds, Alexander Melnikov, Natalia Kliushkina, Carlos Araya, Siqi
  Yan, and Orion Reblitz-Richardson.
\newblock Captum: {A} {Unified} and {Generic} {Model} {Interpretability}
  {Library} for {PyTorch}.
\newblock {\em arXiv:2009.07896 [cs, stat]}, September 2020.
\newblock arXiv: 2009.07896.

\bibitem{krizhevsky_imagenet_2012}
Alex Krizhevsky, Ilya Sutskever, and Geoffrey~E. Hinton.
\newblock {ImageNet} {Classification} with {Deep} {Convolutional} {Neural}
  {Networks}.
\newblock In {\em Proceedings of the 25th {International} {Conference} on
  {Neural} {Information} {Processing} {Systems} - {Volume} 1}, {NIPS}'12, pages
  1097--1105, Red Hook, NY, USA, December 2012. Curran Associates Inc.

\bibitem{lage_evaluation_2018}
Isaac Lage, Emily Chen, Jeffrey He, Menaka Narayanan, Been Kim, Sam Gershman,
  and Finale Doshi-Velez.
\newblock An {Evaluation} of the {Human}-{Interpretability} of {Explanation}.
\newblock In {\em Workshop on {Correcting} and {Critiquing} {Trends} in
  {Machine} {Learning}}, 2018.

\bibitem{lakkaraju_faithful_2019}
Himabindu Lakkaraju, Ece Kamar, Rich Caruana, and Jure Leskovec.
\newblock Faithful and {Customizable} {Explanations} of {Black} {Box} {Models}.
\newblock In {\em Proceedings of the 2019 {AAAI}/{ACM} {Conference} on {AI},
  {Ethics}, and {Society}}, {AIES} '19, pages 131--138, New York, NY, USA,
  January 2019. Association for Computing Machinery.

\bibitem{li_visualizing_2016}
Jiwei Li, Xinlei Chen, Eduard Hovy, and Dan Jurafsky.
\newblock Visualizing and {Understanding} {Neural} {Models} in {NLP}.
\newblock In {\em Proceedings of the 2016 {Conference} of the {North}
  \{{A}\}merican {Chapter} of the {Association} for {Computational}
  {Linguistics}: {Human} {Language} {Technologies}}, January 2016.

\bibitem{lipton_mythos_2018}
Zachary~C. Lipton.
\newblock The {Mythos} of {Model} {Interpretability}.
\newblock {\em Communications of the ACM}, 61(10):36--43, September 2018.

\bibitem{liu_roberta_2019}
Yinhan Liu, Myle Ott, Naman Goyal, Jingfei Du, Mandar Joshi, Danqi Chen, Omer
  Levy, Mike Lewis, Luke Zettlemoyer, and Veselin Stoyanov.
\newblock {RoBERTa}: {A} {Robustly} {Optimized} {BERT} {Pretraining}
  {Approach}.
\newblock {\em arXiv:1907.11692 [cs]}, July 2019.

\bibitem{loshchilov_decoupled_2019}
Ilya Loshchilov and Frank Hutter.
\newblock Decoupled {Weight} {Decay} {Regularization}.
\newblock In {\em International {Conference} on {Learning} {Representations}},
  2019.

\bibitem{lundberg_shap_2018}
Scott Lundberg.
\newblock {SHAP}, 2018.
\newblock https://github.com/slundberg/shap.

\bibitem{lundberg_consistent_2019}
Scott~M. Lundberg, Gabriel~G. Erion, and Su-In Lee.
\newblock Consistent {Individualized} {Feature} {Attribution} for {Tree}
  {Ensembles}.
\newblock {\em arXiv:1802.03888 [cs, stat]}, March 2019.
\newblock arXiv: 1802.03888.

\bibitem{lundberg_unified_2017}
Scott~M. Lundberg and Su-In Lee.
\newblock A {Unified} {Approach} to {Interpreting} {Model} {Predictions}.
\newblock In {\em Proceedings of the 31st {International} {Conference} on
  {Neural} {Information} {Processing} {Systems}}, {NIPS}'17, pages 4768--4777,
  Red Hook, NY, USA, December 2017. Curran Associates Inc.

\bibitem{maas_learning_2011}
Andrew~L. Maas, Raymond~E. Daly, Peter~T. Pham, Dan Huang, Andrew~Y. Ng, and
  Christopher Potts.
\newblock Learning {Word} {Vectors} for {Sentiment} {Analysis}.
\newblock In {\em Proceedings of the 49th {Annual} {Meeting} of the
  {Association} for {Computational} {Linguistics}: {Human} {Language}
  {Technologies} - {Volume} 1}, {HLT} '11, pages 142--150, USA, June 2011.
  Association for Computational Linguistics.

\bibitem{miller_explanation_2019}
Tim Miller.
\newblock Explanation in artificial intelligence: {Insights} from the social
  sciences.
\newblock {\em Artificial Intelligence}, 267:1--38, February 2019.

\bibitem{montavon_layer-wise_2019}
Grégoire Montavon, Alexander Binder, Sebastian Lapuschkin, Wojciech Samek, and
  Klaus-Robert Müller.
\newblock Layer-{Wise} {Relevance} {Propagation}: {An} {Overview}.
\newblock In Wojciech Samek, Grégoire Montavon, Andrea Vedaldi, Lars~Kai
  Hansen, and Klaus-Robert Müller, editors, {\em Explainable {AI}:
  {Interpreting}, {Explaining} and {Visualizing} {Deep} {Learning}}, Lecture
  {Notes} in {Computer} {Science}, pages 193--209. Springer International
  Publishing, Cham, 2019.

\bibitem{paszke_pytorch_2019}
Adam Paszke, Sam Gross, Francisco Massa, Adam Lerer, James Bradbury, Gregory
  Chanan, Trevor Killeen, Zeming Lin, Natalia Gimelshein, Luca Antiga, Alban
  Desmaison, Andreas Kopf, Edward Yang, Zachary DeVito, Martin Raison, Alykhan
  Tejani, Sasank Chilamkurthy, Benoit Steiner, Lu~Fang, Junjie Bai, and Soumith
  Chintala.
\newblock {PyTorch}: {An} {Imperative} {Style}, {High}-{Performance} {Deep}
  {Learning} {Library}.
\newblock {\em Advances in Neural Information Processing Systems},
  32:8026--8037, 2019.

\bibitem{poerner_evaluating_2018}
Nina Poerner, Hinrich Schütze, and Benjamin Roth.
\newblock Evaluating {Neural} {Network} {Explanation} {Methods} using {Hybrid}
  {Documents} and {Morphosyntactic} {Agreement}.
\newblock In {\em Proceedings of the 56th {Annual} {Meeting} of the
  {Association} for {Computational} {Linguistics} ({Volume} 1: {Long}
  {Papers})}, pages 340--350, Melbourne, Australia, July 2018. Association for
  Computational Linguistics.

\bibitem{poursabzi-sangdeh_manipulating_2021}
Forough Poursabzi-Sangdeh, Daniel~G. Goldstein, Jake~M. Hofman,
  Jennifer~Wortman Vaughan, and Hanna Wallach.
\newblock Manipulating and {Measuring} {Model} {Interpretability}.
\newblock In {\em Proceedings of the 2021 {CHI} {Conference} on {Human}
  {Factors} in {Computing} {Systems}}, January 2021.
\newblock arXiv: 1802.07810.

\bibitem{reimers_sentence-bert_2019}
Nils Reimers and Iryna Gurevych.
\newblock Sentence-{BERT}: {Sentence} {Embeddings} using {Siamese}
  {BERT}-{Networks}.
\newblock In {\em Proceedings of the 2019 {Conference} on {Empirical} {Methods}
  in {Natural} {Language} {Processing} and the 9th {International} {Joint}
  {Conference} on {Natural} {Language} {Processing} ({EMNLP}-{IJCNLP})}, August
  2019.

\bibitem{ribeiro_why_2016}
Marco~Tulio Ribeiro, Sameer Singh, and Carlos Guestrin.
\newblock "{Why} {Should} {I} {Trust} {You}?": {Explaining} the {Predictions}
  of {Any} {Classifier}.
\newblock In {\em Proceedings of the 22nd {ACM} {SIGKDD} international
  conference on knowledge discovery and data mining}, August 2016.

\bibitem{ribeiro_anchors_2018}
Marco~Tulio Ribeiro, Sameer Singh, and Carlos Guestrin.
\newblock Anchors: {High}-{Precision} {Model}-{Agnostic} {Explanations}.
\newblock In {\em Thirty-{Second} {AAAI} {Conference} on {Artificial}
  {Intelligence}}, April 2018.

\bibitem{ribeiro_beyond_2020}
Marco~Tulio Ribeiro, Tongshuang Wu, Carlos Guestrin, and Sameer Singh.
\newblock Beyond {Accuracy}: {Behavioral} {Testing} of {NLP} {Models} with
  {CheckList}.
\newblock {\em arXiv:2005.04118 [cs]}, May 2020.

\bibitem{rychener_sentence-based_2020}
Yves Rychener, Xavier Renard, Djamé Seddah, Pascal Frossard, and Marcin
  Detyniecki.
\newblock Sentence-{Based} {Model} {Agnostic} {NLP} {Interpretability}.
\newblock {\em arXiv:2012.13189 [cs, stat]}, December 2020.
\newblock arXiv: 2012.13189.

\bibitem{samek_evaluating_2017}
W.~Samek, A.~Binder, G.~Montavon, S.~Lapuschkin, and K.~Müller.
\newblock Evaluating the {Visualization} of {What} a {Deep} {Neural} {Network}
  {Has} {Learned}.
\newblock {\em IEEE Transactions on Neural Networks and Learning Systems},
  28(11):2660--2673, November 2017.

\bibitem{schmidt_quantifying_2019}
Philipp Schmidt and Felix Biessmann.
\newblock Quantifying {Interpretability} and {Trust} in {Machine} {Learning}
  {Systems}.
\newblock In {\em {AAAI}-19 {Workshop} on {Network} {Interpretability} for
  {Deep} {Learning}}, January 2019.

\bibitem{shrikumar_learning_2017}
Avanti Shrikumar, Peyton Greenside, and Anshul Kundaje.
\newblock Learning {Important} {Features} {Through} {Propagating} {Activation}
  {Differences}.
\newblock In {\em International {Conference} on {Machine} {Learning}}, pages
  3145--3153. PMLR, July 2017.

\bibitem{slack_reliable_2021}
Dylan Slack, {Sophie Hilgard}, {Sameer Singh}, and {Himabindu Lakkaraju}.
\newblock Reliable {Post} hoc {Explanations}: {Modeling} {Uncertainty} in
  {Explainability}.
\newblock In {\em Neural {Information} {Processing} {Systems}}, 2021.

\bibitem{strubell_energy_2019}
Emma Strubell, Ananya Ganesh, and Andrew McCallum.
\newblock Energy and {Policy} {Considerations} for {Deep} {Learning} in {NLP}.
\newblock In {\em Proceedings of the 57th {Annual} {Meeting} of the
  {Association} for {Computational} {Linguistics}}, pages 3645--3650, Florence,
  Italy, July 2019. Association for Computational Linguistics.

\bibitem{sun_how_2019}
Chi Sun, Xipeng Qiu, Yige Xu, and Xuanjing Huang.
\newblock How to {Fine}-{Tune} {BERT} for {Text} {Classification}?
\newblock In Maosong Sun, Xuanjing Huang, Heng Ji, Zhiyuan Liu, and Yang Liu,
  editors, {\em Chinese {Computational} {Linguistics}}, Lecture {Notes} in
  {Computer} {Science}, pages 194--206, Cham, 2019. Springer International
  Publishing.

\bibitem{sundararajan_many_2020}
Mukund Sundararajan and Amir Najmi.
\newblock The {Many} {Shapley} {Values} for {Model} {Explanation}.
\newblock In {\em International {Conference} on {Machine} {Learning}}, pages
  9269--9278. PMLR, November 2020.

\bibitem{sundararajan_axiomatic_2017}
Mukund Sundararajan, Ankur Taly, and Qiqi Yan.
\newblock Axiomatic {Attribution} for {Deep} {Networks}.
\newblock In {\em International {Conference} on {Machine} {Learning}}, June
  2017.

\bibitem{tenney_language_2020}
Ian Tenney, James Wexler, Jasmijn Bastings, Tolga Bolukbasi, Andy Coenen,
  Sebastian Gehrmann, Ellen Jiang, Mahima Pushkarna, Carey Radebaugh, Emily
  Reif, and Ann Yuan.
\newblock The {Language} {Interpretability} {Tool}: {Extensible}, {Interactive}
  {Visualizations} and {Analysis} for {NLP} {Models}.
\newblock {\em arXiv:2008.05122 [cs]}, August 2020.

\bibitem{tjoa_survey_2020}
Erico Tjoa and Cuntai Guan.
\newblock A {Survey} on {Explainable} {Artificial} {Intelligence} ({XAI}):
  {Towards} {Medical} {XAI}.
\newblock {\em IEEE Transactions on Neural Networks and Learning Systems},
  pages 1--21, 2020.
\newblock arXiv: 1907.07374.

\bibitem{wallace_allennlp_2019}
Eric Wallace, Jens Tuyls, Junlin Wang, Sanjay Subramanian, Matt Gardner, and
  Sameer Singh.
\newblock {AllenNLP} {Interpret}: {A} {Framework} for {Explaining}
  {Predictions} of {NLP} {Models}.
\newblock In {\em Proceedings of the 2019 {EMNLP} and the 9th {IJCNLP}
  ({System} {Demonstrations})}, pages 7--12, September 2019.

\bibitem{wang_topical_2007}
X.~Wang, A.~McCallum, and X.~Wei.
\newblock Topical {N}-{Grams}: {Phrase} and {Topic} {Discovery}, with an
  {Application} to {Information} {Retrieval}.
\newblock In {\em Seventh {IEEE} {International} {Conference} on {Data}
  {Mining} ({ICDM} 2007)}, pages 697--702, October 2007.

\bibitem{wolf_transformers_2020}
Thomas Wolf, Lysandre Debut, Victor Sanh, Julien Chaumond, Clement Delangue,
  Anthony Moi, Pierric Cistac, Tim Rault, Remi Louf, Morgan Funtowicz, Joe
  Davison, Sam Shleifer, Patrick von Platen, Clara Ma, Yacine Jernite, Julien
  Plu, Canwen Xu, Teven Le~Scao, Sylvain Gugger, Mariama Drame, Quentin Lhoest,
  and Alexander Rush.
\newblock Transformers: {State}-of-the-{Art} {Natural} {Language} {Processing}.
\newblock In {\em Proceedings of the 2020 {Conference} on {Empirical} {Methods}
  in {Natural} {Language} {Processing}: {System} {Demonstrations}}, pages
  38--45, Online, October 2020. Association for Computational Linguistics.

\bibitem{yang_finbert_2020}
Yi~Yang, Mark Christopher~Siy UY, and Allen Huang.
\newblock {FinBERT}: {A} {Pretrained} {Language} {Model} for {Financial}
  {Communications}.
\newblock {\em arXiv:2006.08097 [cs]}, July 2020.

\bibitem{zafar_lack_2021}
Muhammad~Bilal Zafar, Michele Donini, Dylan Slack, Cedric Archambeau, Sanjiv
  Das, and Krishnaram Kenthapadi.
\newblock On the {Lack} of {Robust} {Interpretability} of {Neural} {Text}
  {Classifiers}.
\newblock In {\em {ACL} {Findings}}, 2021.

\end{thebibliography}
